\documentclass{article} 
\usepackage{iclr2027_conference,times}

\usepackage{amsmath,amsfonts,bm}

\def\eqref#1{equation~\ref{#1}}

\def\1{\bm{1}}

\DeclareMathAlphabet{\mathsfit}{\encodingdefault}{\sfdefault}{m}{sl}
\SetMathAlphabet{\mathsfit}{bold}{\encodingdefault}{\sfdefault}{bx}{n}

\DeclareMathOperator*{\argmax}{arg\,max}
\DeclareMathOperator*{\argmin}{arg\,min}

\usepackage{graphicx}
\usepackage{wrapfig}
\usepackage{hyperref}
\usepackage{url}
\usepackage{booktabs}
\usepackage{tabularx}
\usepackage{colortbl}
\usepackage{amsthm,amsmath,amssymb}
\usepackage{mathrsfs}
\usepackage{booktabs}
\usepackage{graphicx}

\iclrfinalcopy

\title{When Reasoning Goes Astray: Attention Dynamics of Uncontrolled Reasoning}

\author{
 \textbf{Yuanhe Zhang\textsuperscript{1}}, 
 \textbf{Ziwei Wang\textsuperscript{2}}, 
 \textbf{Jie Ren\textsuperscript{1}}, 
 \textbf{Haoran Gao\textsuperscript{3}}, 
 \textbf{Zhenhong Zhou\textsuperscript{4}}, 
 \textbf{Fanyu Meng\textsuperscript{3}},
 \\
 \textbf{Cong Wu\textsuperscript{2}},
 \textbf{Li Sun\textsuperscript{1}},
 \textbf{Sen Su\textsuperscript{1, 5,  $^\dagger$}} 
\\ \textsuperscript{\rm 1}Beijing University of Posts and Telecommunications
\textsuperscript{\rm 3}Wuhan University
\textsuperscript{\rm 3}JIUTIAN Research
\\ \textsuperscript{\rm 4}Nanyang Technological University
\textsuperscript{\rm 5}Chongqing University of Posts and Telecommunications
\\ \{charmes-zhang, susen\}@bupt.edu.cn;
}

\begin{document}

\maketitle
\begingroup
\renewcommand\thefootnote{}\footnotemark
\footnotetext{$\dagger$ indicates corresponding author.}
\endgroup

\begin{abstract}
Large reasoning models (LRMs) improve performance on complex tasks through extended reasoning, yet the same process can degenerate into redundant verification and persistent generation loops.
Such uncontrolled reasoning increases inference cost and creates risks of resource exhaustion and service degradation.
However, existing mitigations largely truncate long outputs or react to surface repetition, and thus fail to distinguish normal thinking from uncontrolled reasoning or explain how benign reasoning degenerates into harmful behavior.
In this paper, we operationalize LRM generation as four states and further introduce \textit{Reasoning-state Analysis via Dynamic Attention Responses} (RADAR), which identifies the current reasoning state in real time and characterizes how effective reflection can develop into uncontrolled generation.
Guided by RADAR's analysis, we further realign abnormal attention distributions toward patterns observed in normal requests and examine how this correction affects excessive reflection and persistent looping.
Temporal analyses show that uncontrolled reasoning is characterized by attention distributions that deviate from normal generation, with abnormal trends becoming detectable before repetition begins. 
Correcting these deviations through Attention Realignment consistently reduces looping while largely preserving benign performance.
Together, RADAR provide a mechanistic account of how reasoning becomes uncontrolled, offering actionable guidance for identifying critical failure stages and designing targeted runtime interventions.
\end{abstract}

\section{Introduction}
\label{sec:introduction}

Large reasoning models (LRMs) improve performance on complex tasks by allocating additional computation to explicit reasoning traces \citep{deepseekai2025deepseekr1,muennighoff-etal-2025-s1, zhang2026resource}.
These traces support reasoning, reflection, and revision of intermediate solutions \citep{deepseekai2025deepseekr1}.
However, models can continue generating redundant solutions after reaching a correct answer, consuming additional tokens with little improvement in accuracy \citep{pmlr-v267-chen25bx,ICLR2026_8df90a14}.
Adversarial inputs can amplify this inefficiency by triggering persistent generation loops until the output budget is exhausted\citep{kumar2025overthink,wang2026recur,li2025loopllm}.
Such prolonged generation increases inference latency and computational cost, and can threaten service availability under limited computing resources \citep{kumar2025overthink,ICLR2025_a815fe7c}.

Existing approaches primarily reduce reasoning cost by shortening or terminating the reasoning process \citep{han-etal-2025-token,hou2025thinkprune}.
Budget based methods set a fixed token budget for reasoning or terminate generation once a predefined limit is reached \citep{han-etal-2025-token,muennighoff-etal-2025-s1}.
Adaptive stopping methods instead monitor prediction confidence during reasoning to determine when further computation is unnecessary \citep{ICLR2026_8df90a14,hosseini2026earlystoppinglargereasoning}.
Other methods compress reasoning traces or optimize the model under explicit length constraints \citep{xia-etal-2025-tokenskip,luo-etal-2026-o1}.
These approaches reduce token consumption, but their accuracy--cost tradeoff depends on the chosen budget or degree of reasoning compression\citep{han-etal-2025-token,xia-etal-2025-tokenskip}.
Moreover, model reasoning plays an important role in solving complex problems, making indiscriminate termination liable to discard useful computation \citep{muennighoff-etal-2025-s1,deepseekai2025deepseekr1}.
We therefore argue that effectively controlling harmful reasoning requires distinguishing reasoning states and analyzing their internal dynamics.

\begin{figure}[t]
    \centering
    \includegraphics[width=\textwidth]{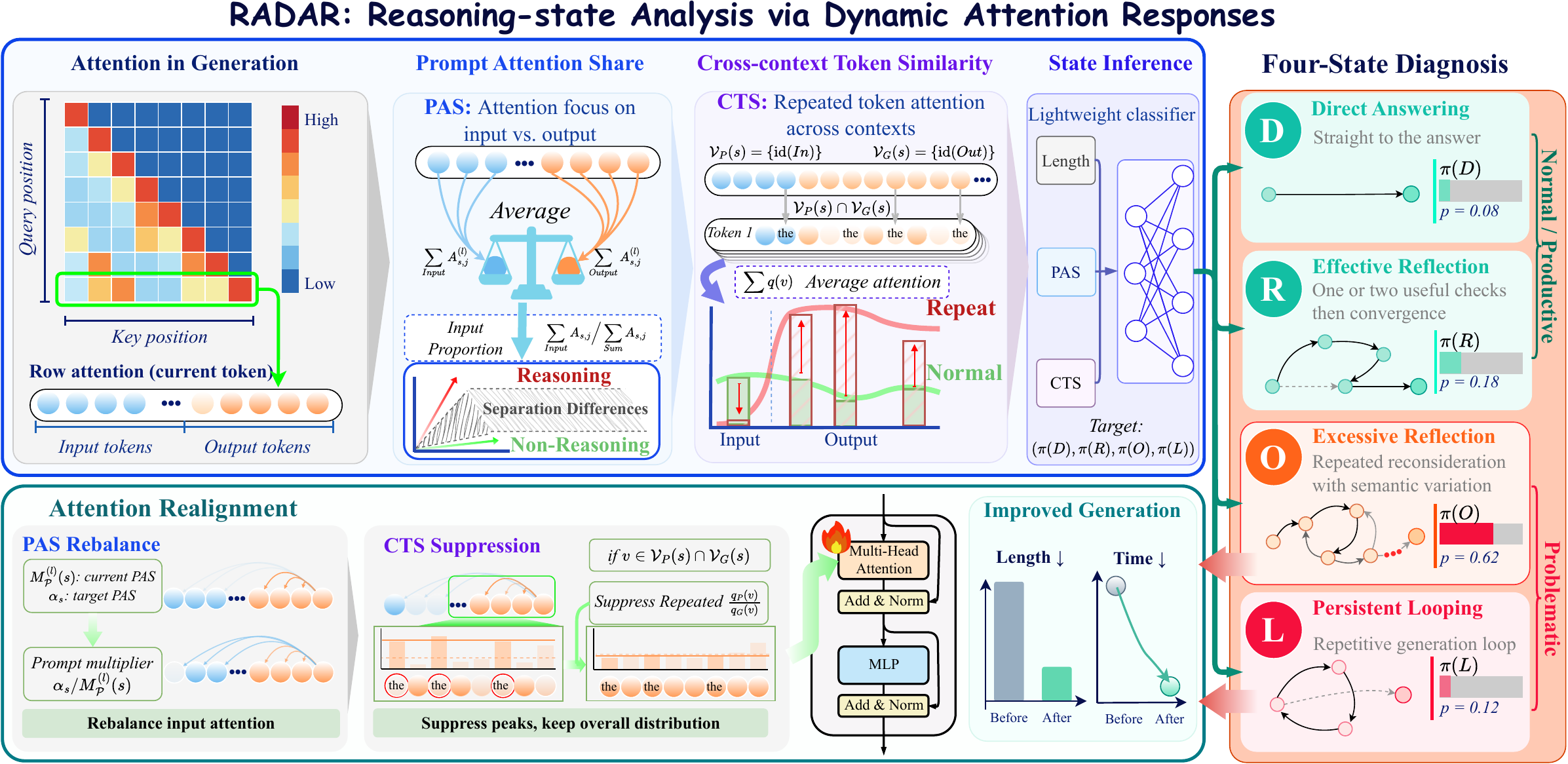}
    \caption{Overview of RADAR, which characterizes attention during generation and supports diagnosis of direct answering, effective reflection, excessive reflection, and persistent looping.}
    \label{fig:radar_overview}
\end{figure}

In this paper, we formulate LRM generation as four Reasoning-States: direct answering ($D$), effective reflection ($R$), excessive reflection ($O$), and persistent looping ($L$).
To identify these reasoning states, we introduce \textit{\textbf{R}easoning-state \textbf{A}nalysis via \textbf{D}ynamic \textbf{A}ttention \textbf{R}esponses} (RADAR), a runtime diagnostic framework that treats attention dynamics as an evolving internal signal of the current reasoning state (Figure~\ref{fig:radar_overview}).
RADAR characterizes these dynamics using two complementary measurements. \textit{Prompt Attention Share} (PAS) quantifies the proportion of attention allocated to the prompt at the current reasoning step, while \textit{Cross-context Token Similarity} (CTS) compares token-level attention distributions between the prompt and generated context.
Using PAS and CTS, a \textit{Temporal State Inference Classifier} encodes their trajectories and estimates the current reasoning state at any stage of generation.
RADAR thereby dynamically evaluates the probabilities of different reasoning tendencies throughout generation and provides a state-aware signal for downstream intervention.
Guided by this signal, we further validate the identified attention patterns through \textit{Attention Realignment}, realigning abnormal attention distributions toward those observed in normal requests and examining the resulting changes in excessive reflection and persistent looping.

We evaluate RADAR across five models. RADAR achieves an average micro-F1 of 89.8\% for Reasoning-States diagnosis. These abnormal trends become detectable before the recorded onset of repetition in naturally induced attacks. 
Guided by these observations, Attention Realignment reduces the average loop rate by 9.8 percentage points across five models while largely preserving benign performance. Together, these results show that attention dynamics provide a reliable signal for state diagnosis, transition analysis, and targeted runtime intervention without treating all extended reasoning as harmful.

Overall, we introduce RADAR to characterize and track four functional reasoning states through evolving attention dynamics, revealing how effective reflection can develop into excessive reflection and persistent looping. 
Building on these findings, we further develop a state guided intervention to examine and modulate abnormal reasoning dynamics while preserving normal reasoning.

\section{Related Work}
\label{sec:related_work}
\subsection{Performance of Large Reasoning Models}
\label{sec:related_lrm_performance}

Early work showed that explicit reasoning can substantially improve language model performance by exposing and refining intermediate solution steps. Chain of thought prompting enables few shot and zero shot reasoning \citep{NEURIPS2022_9d560961,NEURIPS2022_8bb0d291}, while self consistency and tree search further improve solution quality by exploring multiple reasoning paths \citep{Wang2023,NEURIPS2023_271db992}. Outcome and process verifiers complement these approaches by providing answer level and step level supervision for selecting more reliable solutions \citep{cobbe2021training,ICLR2024_aca97732}. More recent large reasoning models (LRMs) move beyond eliciting reasoning primarily through prompting and instead acquire stronger reasoning capabilities through dedicated training, including rationale self training and reinforcement learning, as demonstrated by DeepSeek R1, and QwQ \citep{NEURIPS2022_639a9a17,shao2024deepseekmathpushinglimitsmathematical,deepseekai2025deepseekr1,qwen2025qwq}. Building on these advances, current models and inference strategies further exploit test time computation through thinking modes, budget control, adaptive compute allocation, and budget forcing \citep{yang2025qwen3,ICLR2025_1b623663,muennighoff-etal-2025-s1}. Together, these developments establish explicit reasoning as key sources of LRM performance, motivating the need to preserve useful reasoning rather than treating extended reasoning traces as inherently harmful.

\subsection{Resource Consumption and Defenses}
\label{sec:related_resources_defenses}
Additional reasoning does not guarantee additional utility. Overthinking can produce redundant solutions \citep{pmlr-v267-chen25bx}, missing premises can prolong ineffective deliberation \mbox{\citep{fan2025missing}}, and longer reasoning can even reduce accuracy under distractibility and spurious correlations \citep{b6ae0c676d5c4d9cac57d8cf2ece527e}. Such inefficiency can be deliberately amplified by availability attacks. Sponge examples increase energy consumption and latency \citep{128fcaa8f9204f7d9d4698f555ead185}, Engorgio suppresses termination through adversarial prompts \citep{ICLR2025_a815fe7c}, and P-DoS induces prolonged repetition through training-data poisoning \citep{gao2024denialofservicepoisoningattackslarge}. In reasoning systems, external context attacks can further disrupt deliberation by introducing distracting computations or contradictions \citep{kumar2025overthink,zhang2026codecontradictionbaseddeliberationextension}, while RECUR exploits counterfactual reflection \citep{wang2026recur} and LoopLLM induces repetitive decoding loops \citep{li2025loopllm}. These distinct failure mechanisms motivate separating excessive reflection from persistent lexical looping.

Existing mitigations reduce reasoning cost through budget control, token compression, adaptive stopping, or training for more concise reasoning \citep{han-etal-2025-token,xia-etal-2025-tokenskip,ICLR2026_8df90a14,hosseini2026earlystoppinglargereasoning,hou2025thinkprune,luo-etal-2026-o1,NEURIPS2025_44f09d19, zhang2025pd3f}. Attention intervention has also been explored, for example by biasing output attention to suppress continued reasoning after an injected end-of-thinking token \citep{koh2026thinkdoesntstopreasoning}. 
However, these methods do not explicitly identify the functional state of ongoing reasoning, particularly whether it has become excessively reflective or persistently repetitive.
RADAR instead models prompt--generation attention dynamics to identify four Reasoning-States and guide uncontrolled reasoning intervention, while evaluating both attack suppression and preservation of useful reflection.

\section{Method}
\label{sec:method}

Our method contains three parts. Section~\ref{sec:method_threat_model} specifies the threat model and defines four functional reasoning states. Section~\ref{sec:method_radar} identifies the current reasoning state during generation by constructing a Temporal State Inference Classifier from PAS and CTS. Section~\ref{sec:method_radar_d} presents Attention Realignment, which adjusts abnormal attention patterns toward those observed in normal reasoning to examine their role in excessive reflection and persistent looping.
Figure~\ref{fig:radar_overview} summarizes the diagnostic pipeline and the state-guided defense.

\subsection{Threat Model and Reasoning States}
\label{sec:method_threat_model}

We consider an LRM exposed through a public inference service.
Each request is processed under a finite generation budget, but the amount of this budget consumed by an individual response is not known in advance.
Uncontrolled reasoning, whether it arises spontaneously or is induced by an adversarial input, can sustain excessive reflection or persistent looping until the budget is exhausted.
This behavior increases inference cost and may substantially delay natural termination.
Under this threat model, we assume a service-side defender that can access and modify the model's attention tensors during generation to support runtime mitigation.
Although uncontrolled reasoning generally produces redundant tokens, this shared outcome can arise from different failure mechanisms, limiting the effectiveness of one size fits all mitigation.
A finer distinction is therefore necessary to disentangle these behaviors and reveal their underlying mechanisms.
Let the input request be $\mathbf{x}=(x_1,\ldots,x_p)$ and a complete generation be $\mathbf{y}=(y_1,\ldots,y_t)$.
Their concatenation forms a sequence $\mathbf{u}$ of length $s=p+t$.
Following common distinctions between reflective reasoning and semantic cycling, we organize generation trajectories into four Reasoning-States:
\begin{itemize}
    \item \textbf{Direct answering ($D$).} The model produces a definite final answer without explicit reflection or backtracking. Answer correctness does not affect this label.
    \item \textbf{Effective reflection ($R$).} The model uses a bounded sequence of correction or reconsideration steps that contributes to the solution and ultimately produces a definite final answer.
    \item \textbf{Excessive reflection ($O$).} The model continues reconsidering after useful progress has saturated, repeatedly revisits semantically similar issues, and contributes little new information toward a final answer.
    \item \textbf{Persistent looping ($L$).} The model repeatedly emits identical content, makes no substantive progress, and does not terminate naturally within the available budget.
\end{itemize}
We denote the resulting state space as $\mathcal{S}=\{D,R,O,L\}$.
Appendix~\ref{app:state_annotation} provides representative log examples and annotation criteria for the four reasoning states.

\subsection{Temporal State Inference with RADAR}
\label{sec:method_radar}
To distinguish these Reasoning-State during generation, we propose RADAR, a reasoning state classifier based on attention dynamics in LRMs. 
We begin by representing the layerwise attention patterns used to construct PAS and CTS. At sequence length $s$, we collect the causal attention matrices from all $L$ transformer layers as $\mathcal{A}_s=\{\mathbf{A}^{(l)}_s\in[0,1]^{s\times s}\}_{l=1}^{L}$, where $A^{(l)}_{i,j}$ denotes the attention weight from position $i$ to position $j$.

\paragraph{Prompt Attention Share (PAS)}
\label{sec:method_pas}
characterizes how the model partitions attention between the original prompt and the generated context at the current reasoning step. 
Under causal attention, the last row of each layerwise attention matrix represents the current token's attention distribution at that step.
We define PAS at sequence length $s$ by first measuring, at each layer, the fraction of attention assigned to the positions $p$ and then averaging this fraction across all $L$ layers:
\begin{equation}
M_\mathcal{P}(s)
=
\mathbb{E}_{l\in\{1,\ldots,L\}}
\left[
\frac{\sum_{j=p}^{s} A_{s,j}^{(l)}}{\sum_{j=1}^{s} A_{s,j}^{(l)}}
\right]
=
\mathbb{E}_{l}
\left[
\sum_{j=p}^{s} A_{s,j}^{(l)}
\right].
\end{equation}
Since each attention row is normalized, the denominator satisfies $\sum_{j=1}^{s} A_{s,j}^{(l)}=1$. We retain the ratio form to explicitly represent the relative allocation of attention to the original prompt. Higher PAS indicates greater attention to the prompt, whereas lower PAS reflects a shift toward the generation.

\paragraph{Cross-context Token Similarity (CTS)}
\label{sec:method_cts}
compares the attention patterns associated with the same token identity. At sequence length $s$, we partition the token positions into the prompt region $\mathcal{P}=\{1,\ldots,p\}$ and the generation region $\mathcal{G}=\{p+1,\ldots,s\}$. We use $\operatorname{id}(j)$ to map each position $j$ to its original token.
To describe both regions uniformly, let $X\in\{P,G\}$ and define $\mathcal{I}_X(s)=\mathcal{X}$. The set $\mathcal{V}_X(s)=\{\operatorname{id}(j):j\in\mathcal{I}_X(s)\}$ contains the distinct token identities observed in region $X$. Thus, a token identity appears once in $\mathcal{V}_X(s)$ even if it occurs at multiple positions in $\mathcal{I}_X(s)$.

For layer $l$ and token identity $v\in\mathcal{V}_P(s)\cap\mathcal{V}_G(s)$, we aggregate the current token's attention over every occurrence of $v$ in region $X$ and normalize it by the total attention assigned to that region:
\begin{equation}
    q_X^{(l)}(v)=
    \frac{\sum_{j\in\mathcal{I}_X(s)}\mathbb{I}[\operatorname{id}(j)=v]\,A_{s,j}^{(l)}}
    {\sum_{j\in\mathcal{I}_X(s)}A_{s,j}^{(l)}}.
\end{equation}
The resulting $q_X^{(l)}(v)$ is the within-region attention share of identity $v$. 
We compute the total-variation similarity at each layer and average it across layers:
\begin{equation}
    M_\mathcal{C}(s)=\frac{1}{L}\sum_{l=1}^{L}\left[
    1-\frac{1}{2}
    \sum_{v\in\mathcal{V}_P(s)\cap\mathcal{V}_G(s)}
    \left|q_P^{(l)}(v)-q_G^{(l)}(v)\right|\right].
\end{equation}
Absent token identities receive zero probability in the corresponding region.
Thus, $M_C(s)\in[0,1]$, with higher values indicating more similar token-level attention distributions across the two regions.

\paragraph{Temporal State Inference Classifier.}
\label{sec:method_classifier}
The third RADAR module summarizes the temporal evolution of PAS and CTS into compact features and uses the offline Temporal State Inference Classifier to estimate probabilities over the four reasoning states.
For each training sample $i\in\{1,\ldots,N\}$, let $p_i$ be its prompt length and $k_i$ its generated prefix length. We extract three features from the corresponding PAS and CTS histories, evaluated at total sequence positions $p_i+k$ for $1\leq k\leq k_i$; the sample index on these measurements is omitted for readability.
The first feature captures the temporal trend of PAS by fitting a least-squares line to its history:
\begin{equation}
    (\hat{\beta}_{0,i},\hat{\beta}_{1,i})
    =
    \argmin_{\beta_0,\beta_1}
    \sum_{k=1}^{k_i}
    \left(
        M_{\mathcal P}(p_i+k)-\beta_0-\beta_1 k
    \right)^2,
    \qquad
    f_{i,1}=\hat{\beta}_{1,i},
\end{equation}

The second feature summarizes the overall CTS level up to position $k_i$ as
$f_{i,2}=\operatorname{mean}_{1\leq k\leq k_i} M_{\mathcal C}(p_i+k)$,
capturing the average cross-context similarity throughout the current generation prefix.
The third feature encodes the current generation progress as $f_{i,3}=\log k_i$.


Together, these features form the trajectory representation
$\mathbf{f}_i=[f_{i,1},f_{i,2},f_{i,3}]^{\top}\in\mathbb{R}^{3}$.
The offline training set is defined as
$\mathcal{D}_{\mathrm{train}}=\{(\mathbf{f}_i,c_i)\}_{i=1}^{N}$,
with $c_i\in\mathcal{S}$ denoting the Reasoning-State label of the $i$-th generation prefix.
Prefixes extracted at different positions from the same generation are treated as distinct training samples.
Before classifier training, we compute the training-set mean $\mu_m$ and standard deviation $\sigma_m$
for each feature dimension $m\in\{1,2,3\}$ and standardize it as
$z_{i,m}=(f_{i,m}-\mu_m)/\sigma_m$.
The resulting standardized feature vector is
$\mathbf{z}_i=[z_{i,1},z_{i,2},z_{i,3}]^{\top}\in\mathbb{R}^{3}$.
The Temporal State Inference Classifier estimates the probability of each Reasoning-State as:
\begin{equation}
    \pi_c(s)=\frac{\exp\!\left(\mathbf{w}_c^{\top}\mathbf{z}(s)+b_c\right)}{\sum_{c'\in\mathcal{S}}\exp\!\left(\mathbf{w}_{c'}^{\top}\mathbf{z}(s)+b_{c'}\right)},
    \quad c\in\mathcal{S}.
\end{equation}
where $\mathbf{w}_c\in\mathbb{R}^{3}$ and $b_c$ are the weight vector and bias associated with state $c$.
Let
$\mathbf{W}=[\mathbf{w}_c^{\top}]\in\mathbb{R}^{4\times 3}$
and
$\mathbf{b}=(b_c)\in\mathbb{R}^{4}$
collect the classifier parameters.
To account for class imbalance, we train the classifier with a weighted negative log likelihood objective and $L_2$ regularization.
Let $N_{c_i}$ denote the number of training samples whose Reasoning-State label is $c_i$, and let $\lambda$ control the regularization strength.
The training objective is:
\begin{equation}
\mathcal{L}(\mathbf{W},\mathbf{b})
=
-\sum_{i=1}^{N}
\frac{1}{|\mathcal{S}|N_{c_i}}
\log \pi_{c_i}(i)
+
\frac{\lambda}{2}
\left\|\mathbf{W}\right\|_{F}^{2}.
\end{equation}

After training, the statistics used for feature standardization, together with $\mathbf{W}$ and $\mathbf{b}$.
The state distribution and predicted label are $\boldsymbol{\pi}(s)=\left(\pi_D(s),\pi_R(s),\pi_O(s),\pi_L(s)\right)$, and the current reasoning state is predicted as $\hat{c}(s)=\argmax_{c\in\mathcal{S}}\pi_c(s)$.

\subsection{Attention Realignment in RADAR}
\label{sec:method_radar_d}

Attention Realignment uses the classifier outputs defined above to selectively modulate attention and mitigate uncontrolled reasoning.
During inference, RADAR performs state inference at geometrically spaced sequence lengths
$\mathcal{K}=\{2^n\}_{n=0}^{\lfloor\log_2 S_{\max}\rfloor}$,
where $S_{\max}$ denotes the model's maximum context length.
At each checkpoint $s\in\mathcal{K}$ reached during generation ($s>p$), RADAR computes the state distribution $\boldsymbol{\pi}(s)$ and predicted label $\hat{c}(s)$.
Given a confidence threshold $\tau\in(0,1)$, Attention Realignment is activated when
$\hat{c}(s)\in\{O,L\}$ and $\pi_{\hat{c}(s)}(s)>\tau$.
Once activated, the intervention starts from the next decoding step and remains active until reasoning terminates.

For each layer $l$, we denote its prompt attention share and cross-context token similarity
at sequence length $s$ by $M_{\mathcal P}^{(l)}(s)$ and $M_{\mathcal C}^{(l)}(s)$,
respectively, defined analogously to the layer-averaged PAS and CTS in
Section~\ref{sec:method_classifier}.
We compute the training-set average layerwise PAS as
$\overline{M}_{\mathcal P}^{(l)}
=\operatorname{mean}_{i=1}^{N} M_{\mathcal P,i}^{(l)}(s_i)$
and retain the $\lceil\xi L\rceil$ layers with the largest values, forming the static
candidate set
$\mathcal{J}_{\mathrm{base}}
=\operatorname{Top}_{\lceil\xi L\rceil}
\{\overline{M}_{\mathcal P}^{(l)}\}_{l=1}^{L}$,
with $\xi\in(0,1]$ controlling the retained layer fraction.
This restriction bounds the runtime cost of subsequent layer-wise analysis and intervention.

At each active decoding step, we further retain only layers whose current CTS falls below $\kappa_{\mathcal C}$:
\begin{equation}
    \mathcal{J}_s
    =\left\{l\in\mathcal{J}_{\mathrm{base}}:
    M_{\mathcal C}^{(l)}(s)<\kappa_{\mathcal C}\right\}.
    \label{eq:radar_d_dynamic_layers}
\end{equation}
Thus, intervention is restricted to high-PAS layers that simultaneously exhibit abnormal prompt--generation attention similarity.

\paragraph{PAS rebalance.}
We estimate the normal prompt-attention pattern from training trajectories labeled as direct answering or effective reflection, denoted by $\mathcal{T}_{DR}$. Let $t=s-p$ denote the generated length.
Because PAS is inherently affected by sequence length, with uniform attention corresponding to $M_{\mathcal P}(s)=m_{\mathcal P}\cdot (t/s)$, we normalize $m_{\mathcal P}$ by this baseline and model the ratio $sM_{\mathcal P}(s)/t$.  We choose the basis expansion $\boldsymbol{\phi}(t,p)=(1,\log(t+1),[\log(t+1)]^2,\log p)^{\top}$ and estimate the reference parameters by least squares in the log domain
$\hat{\boldsymbol{\theta}}=\argmin_{\boldsymbol{\theta}}\left\langle\left[\log\!\left(\frac{s}{t}M_{\mathcal P}(s)\right)-\boldsymbol{\theta}^{\top}\boldsymbol{\phi}(t,p)\right]^2\right\rangle_{\mathcal{T}_{DR}}$.

At runtime, the fitted reference is mapped back to the original PAS scale:
\begin{equation}
\alpha_s=\operatorname{clip}_{[\epsilon,1-\epsilon]}\left[\frac{p}{s}\exp\!\left(\hat{\boldsymbol{\theta}}^{\top}\boldsymbol{\phi}(t,p)\right)\right].
\end{equation}

\paragraph{CTS suppression.}
For each selected layer $l \in \mathcal{J}_s$, we compare the $q_P^{(l)}(v)$ and $q_G^{(l)}(v)$ defined in Section~\ref{sec:method_cts}. 
We collect all token identities satisfying $q_G^{(l)}(v)>q_P^{(l)}(v)$ in the suppression set $\mathcal{V}_{\downarrow}^{(l)}$.
We reduce their generation-side attention shares toward the corresponding prompt-side shares and redistribute the released mass over the remaining generation tokens:
\begin{equation}
\begin{aligned}
    \gamma_v^{(l)}
    =\frac{q_P^{(l)}(v)}{q_G^{(l)}(v)},\quad\quad
    \gamma_{\mathrm{rem}}^{(l)}
    =\frac{1-\sum_{v\in\mathcal{V}_{\downarrow}^{(l)}}q_P^{(l)}(v)}
            {1-\sum_{v\in\mathcal{V}_{\downarrow}^{(l)}}q_G^{(l)}(v)},\quad\quad
    v\in\mathcal{V}_{\downarrow}^{(l)}.
\end{aligned}
\end{equation}
Here, $\gamma_v^{(l)}$ suppresses the selected token identities, while $\gamma_{\mathrm{rem}}^{(l)}$ preserves the total generation-side attention mass by reallocating the released share to the remaining tokens.
We then realign the attention distribution toward that observed in normal requests, with the resulting modulation $\zeta_{s,j}^{(l)}$ applied to the attention weights at each selected layer $l$.
\begin{equation}
    \zeta_{s,j}^{(l)}=
    \begin{cases}
        \displaystyle\frac{\alpha_s}{M_{\mathcal P}^{(l)}(s)},
            &j\in\mathcal{P},\\[6pt]
        \displaystyle\frac{1-\alpha_s}{1-M_{\mathcal P}^{(l)}(s)}
            \gamma_{\operatorname{id}(j)}^{(l)},
            &j\in\mathcal{G},\ \operatorname{id}(j)\in\mathcal{V}_{\downarrow}^{(l)},\\[6pt]
        \displaystyle\frac{1-\alpha_s}{1-M_{\mathcal P}^{(l)}(s)}
            \gamma_{\mathrm{rem}}^{(l)},
            &j\in\mathcal{G},\ \operatorname{id}(j)\notin\mathcal{V}_{\downarrow}^{(l)}.
    \end{cases}
    \label{eq:radar_d_multiplier}
\end{equation}

\section{Experiments}
\label{sec:experiments}

\subsection{Experimental Setup}
\label{sec:exp_setup}

\paragraph{Models.}
We evaluate five reasoning models with accessible attention tensors: DeepSeek-R1-Distill-Llama-8B, DeepSeek-R1-Distill-Qwen-14B~\citep{deepseekai2025deepseekr1}, QwQ-32B~\citep{qwen2025qwq}, Qwen-3.6-27B~\citep{yang2025qwen3}, and GLM-4.7-Flash~\citep{zai2026glm47}. 

\paragraph{Tasks and Uncontrolled Reasoning Samples.}
We construct reasoning trajectories covering four Reasoning-States. 
For trajectories expected to admit direct answers ($D$), we use GSM8K~\citep{cobbe2021training} and MMLU-Geor~\citep{hendrycks2021measuring}.
For trajectories requiring more deliberate reasoning ($R$), we use GPQA~\citep{rein2023gpqa} together with MMLU-History and MMLU-Econometrics.
The uncontrolled-reasoning sources ($O\&L$) are Recur~\citep{wang2026recur}, LoopLLM~\citep{li2025loopllm}, Joint construction that concatenates repetition-inducing outputs, and Missing Premise (MiP)~\citep{fan2025missing}. Among them, Recur targets excessive reflection ($O$).

\paragraph{Metrics and Reproducibility.}
For four-state diagnosis, we report micro-F1, the harmonic mean of micro-averaged precision and recall; in our single-label multiclass setting, it is equivalent to overall accuracy~\citep{sokolova2009systematic}.
Detection and intervention analyses use generation length, loop rate, and task accuracy.
Full generation settings, evaluation cohorts, and ablation protocols are provided in Appendices~\ref{app:experimental_settings}, \ref{app:defense_comparison}, and \ref{app:ablation}.

\subsection{Classification Accuracy}
\label{sec:exp_detection}

\paragraph{Four-State Detection Across Models.}
\begin{figure*}[t]
    \centering
    \includegraphics[width=\textwidth]{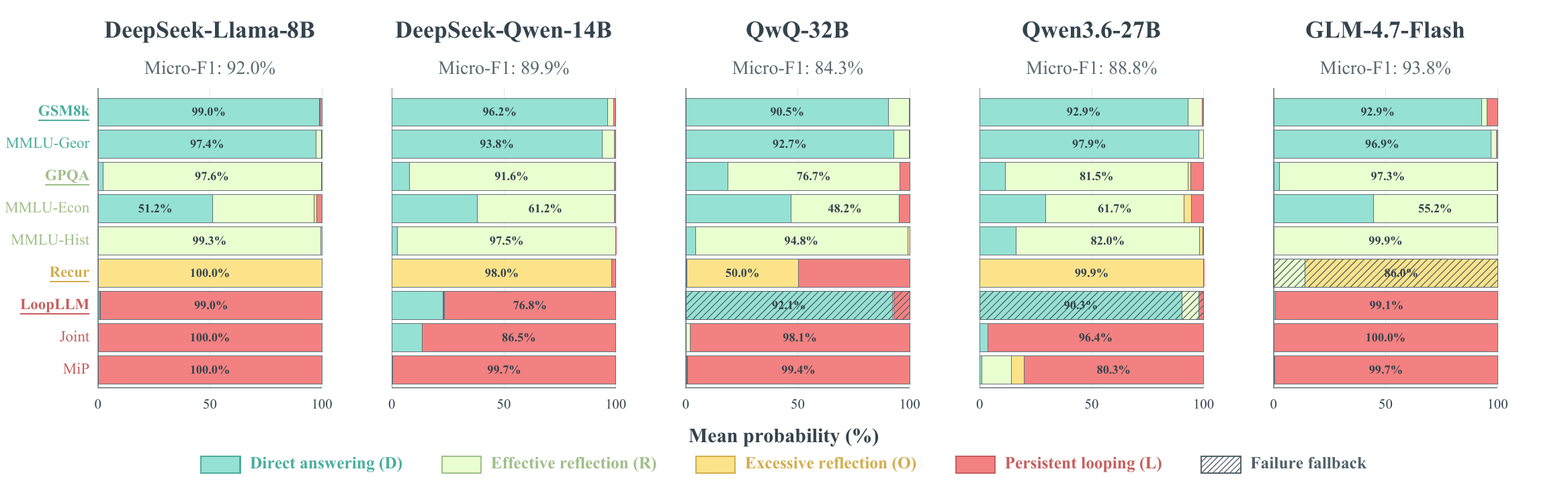}
    \caption{
    Four-state confidence distributions across five models and nine test datasets. Underlined dataset names indicate datasets from which training samples are drawn, while hatched bars indicate model–dataset pairs for which no attack succeeds.
    }
    \label{fig:four_state_accuracy}
\end{figure*}
Figure~\ref{fig:four_state_accuracy} demonstrates the effectiveness of RADAR in distinguishing the four reasoning states across different model families and datasets. The micro-F1 scores with an average of 89.8\%, indicating consistently strong state classification performance across all five models. 
RADAR nevertheless maintains clear state separation on these unseen datasets and attack settings, demonstrating substantial cross-dataset and cross-attack generalization. 

\paragraph{Detection Timing Across Generation.}
\begin{figure*}[t]
    \centering
    \includegraphics[width=\textwidth]{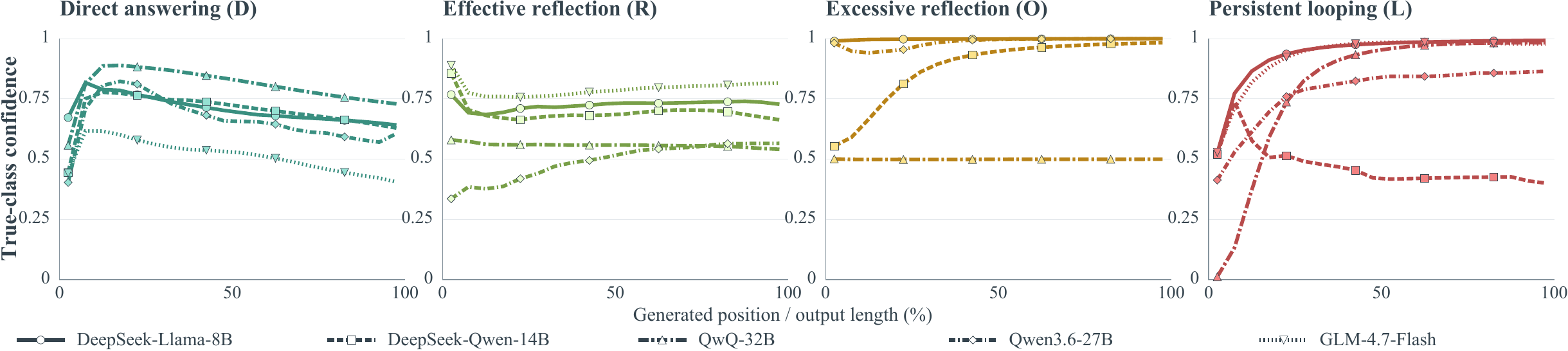}
    \caption{
    Ground-truth class confidence over relative generation progress. Confidence is aggregated over 5\%-wide progress bins, with trajectory-level values averaged equally across samples. 
    For the O and L states, only successful attack trajectories are included.
    }
    \label{fig:detection_accuracy_by_length}
\end{figure*}
Figure~\ref{fig:detection_accuracy_by_length} shows that the four reasoning states exhibit distinct confidence trajectories that are broadly consistent across model families. 
For direct answering, confidence rises rapidly and then stabilizes. By contrast, effective reflection starts with relatively high confidence, suggesting that the tendency to engage in reflection is already evident early in generation. 
This is consistent with prior findings that hidden states can encode properties of future model behavior before the final answer is formed \citep{zhang2025reasoningmodelsknowtheyre}. 
Excessive reflection is recognized with high confidence from early generation and remains stable, whereas Persistent looping emerges progressively. 
Although these curves use trajectory-level rather than token-level labels, they show that state evidence becomes clearly distinguishable after only a short initial stage of generation.



\subsection{Mechanism Analysis}
\label{sec:exp_mechanism}

\paragraph{Attention Signatures.}
\begin{figure}[t]
    \centering
    \vspace{-20pt}
    \includegraphics[width=\linewidth]{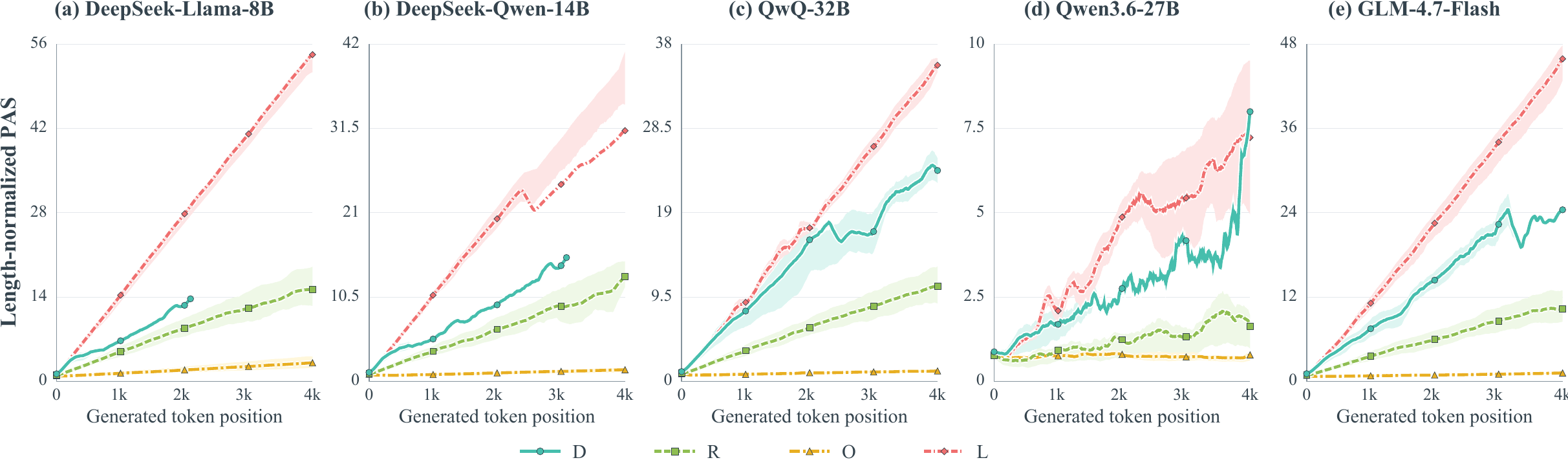}
    \caption{
    Mean length-normalized PAS across generated positions for five models. Solid lines denote the mean PAS for each reasoning state, while shaded regions show the interquartile range across trajectories. Trajectories are truncated at generated position 4095 for visualization.
    }
    \label{fig:pas_slope_distribution}
\end{figure}
PAS measures how attention is allocated between the original prompt and the generated context. 
Figure~\ref{fig:pas_slope_distribution} shows the evolution of normalized PAS over generated positions, reflecting how strongly attention is anchored to the generated context.
For the two benign states, both direct answering $D$ and  effective reflection $R$ increase with generation progress, but $D$ consistently maintains a higher normalized PAS than $R$. This suggests that direct answering remains more strongly anchored to the generation, whereas effective reflection shifts relatively more attention toward the prompt when revising or verifying intermediate steps. 
The two uncontrolled states exhibit substantially different PAS dynamics. Excessive reflection $O$ remains relatively high and stable, indicating that the model continues to anchor strongly to the input prompt throughout prolonged reflection. In contrast, Persistent looping $L$ shows a progressive shift toward the recently generated context, suggesting that the model increasingly anchors on local output patterns as the loop develops.
Importantly, the early overlap between $D$ and $L$ shows that loop trajectories can initially resemble normal generation, whereas their later divergence provides a characteristic temporal signature of loop formation. 

\paragraph{Transitions from Effective to Harmful Reasoning.}
\label{sec:exp_transitions}
\begin{table*}[t]
\centering
\small
\setlength{\tabcolsep}{7pt}
\renewcommand{\arraystretch}{1.20}
\caption{Detection confidence over generation progress. Results include only successful attacks. Entries are mean zero-based generation positions, where $t_{\mathrm{loop}}$ denotes the recorded onset of repetition and $t_\tau$ the earliest position satisfying $P(O)+P(L)\geq\tau$.}
\label{tab:detection_timeliness}
\rowcolors{2}{gray!10}{white}
\begin{tabular}{l|c|cccccc}
\toprule
\textbf{Attack} & $t_{\mathrm{loop}}$ & $t_{0.4}$ & $t_{0.5}$ & $t_{0.6}$ & $t_{0.7}$ & $t_{0.8}$ & $t_{0.9}$ \\
\midrule
Recur & 994.3 & 62.3 & 75.1 & 108.0 & 266.9 & 1052.0 & 1911.3 \\
LoopLLM & 2141.0 & 801.0 & 997.0 & 1265.0 & 1268.7 & 1676.7 & 2461.3 \\
Joint & 0 & 52.2 & 60.6 & 69.2 & 88.1 & 132.8 & 204.7 \\
MiP & 4447.3 & 367.8 & 450.4 & 578.4 & 734.8 & 1004.0 & 1598.8 \\
\bottomrule
\end{tabular}
\end{table*}

\begin{wrapfigure}{r}{0.50\textwidth}
    \vspace{-20pt}
    \centering
    \includegraphics[width=\linewidth]{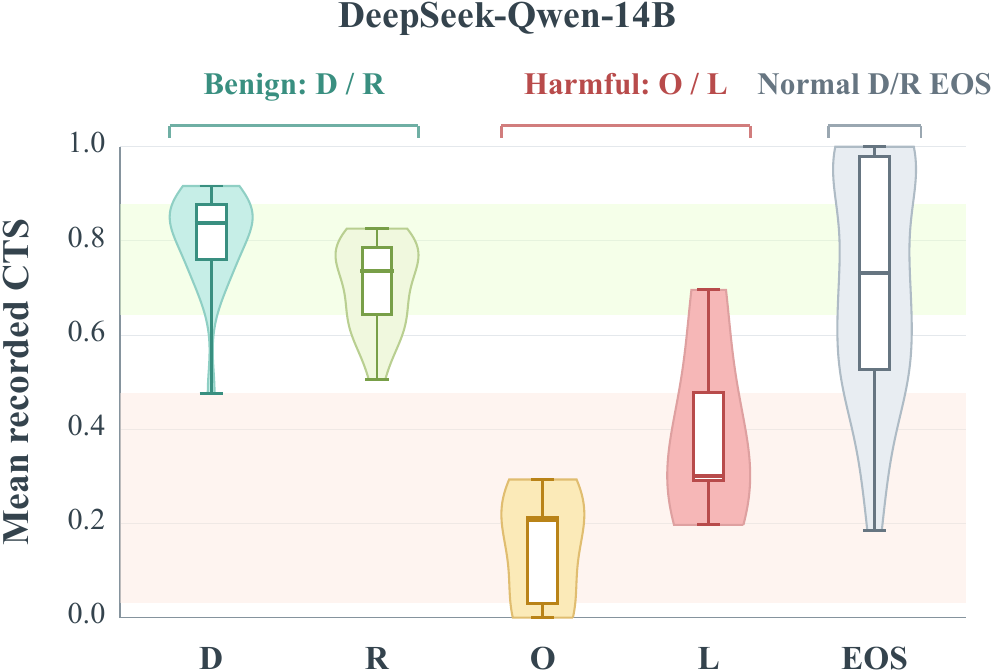}
    \vspace{-9pt}
    \caption{
    CTS distributions on DeepSeek-Qwen.
    }
    \label{fig:cts_distribution}
\end{wrapfigure}
Table~\ref{tab:detection_timeliness} converts the progressive confidence pattern into an operational detection timeline. Across successful attacks, moderate harmful thresholds are reached substantially earlier than very high confidence thresholds. 
More importantly, for all naturally induced attack groups, thresholds up to 70\% are crossed before the recorded onset of repetition; Joint is excluded from this comparison because it directly manipulates the output process. This indicates that harmful-state confidence generally strengthens as generation proceeds, but waiting for 80--90\% confidence can postpone intervention until close to or even after the observed failure onset. 
In contrast, thresholds at or below 50\% provide a more practical operating range for early detection, allowing RADAR to signal harmful behavior before persistent looping emerges.


\paragraph{Mechanism Transfer.}
\label{sec:exp_mechanism_transfer}
CTS measures the consistency of attention assigned to token identities shared by the prompt and generated context. Figure~\ref{fig:cts_distribution} shows that benign states maintain substantially higher CTS than uncontrolled states, with mean values of 0.81 and 0.71 for $D$ and $R$, compared with 0.12 and 0.39 for $O$ and $L$. This suggests that uncontrolled reasoning is associated with stronger prompt--generation attention asymmetry. The same overall trend is observed across all five models.


\subsection{Intervention Effectiveness across Models}
\label{sec:exp_defense}

We use attention realignment as a diagnostic intervention that connects the mechanistic analysis to observable behavior. RADAR corrects abnormal attention allocation by moving PAS and CTS toward the ranges observed in normal trajectories. If the identified attention patterns participate in model collapse, this correction
should reduce persistent looping while leaving benign behavior largely unchanged. We test these two predictions jointly across five model families.
\begin{table*}[t]
    \centering
    \small
    \vspace{-15pt}
    \setlength{\tabcolsep}{5pt}
    \renewcommand{\arraystretch}{1.15}
    \caption{Cross-model effectiveness of Attention Realignment. $\Delta$ is computed as RADAR minus Original. Negative values indicate improvements in final loop rate, whereas positive values indicate improvements in benign accuracy.}
    \label{tab:utility_preservation}
    \begin{tabularx}{\textwidth}{>{\raggedright\arraybackslash}Xcccccc}
        \toprule
        & \multicolumn{3}{c}{\textbf{Final loop rate $\downarrow$}} & \multicolumn{3}{c}{\textbf{Benign accuracy $\uparrow$}} \\
        \cmidrule(lr){2-4}\cmidrule(lr){5-7}
        \textbf{Model} & \textbf{Original} & \textbf{RADAR} & \textbf{$\Delta$ (pp)} & \textbf{Original} & \textbf{RADAR} & \textbf{$\Delta$ (pp)} \\
        \midrule
        \rowcolor[HTML]{F3F4F6} DeepSeek-R1-Distill-Llama-8B & 56.0\% & 40.0\% & \textbf{$-16.0$} & 66.4\% & 65.6\% & \textbf{$-0.8$} \\
        DeepSeek-R1-Distill-Qwen-14B & 40.8\% & 24.8\% & \textbf{$-16.0$} & 72.8\% & 74.4\% & \textbf{$+1.6$} \\
        \rowcolor[HTML]{F3F4F6} QwQ-32B & 22.4\% & 18.4\% & \textbf{$-4.0$} & 71.2\% & 76.0\% & \textbf{$+4.8$} \\
        GLM-4.7-Flash & 36.8\% & 26.4\% & \textbf{$-10.4$} & 65.6\% & 69.6\% & \textbf{$+4.0$} \\
        \rowcolor[HTML]{F3F4F6} Qwen3.6-27B & 28.8\% & 26.4\% & \textbf{$-2.4$} & 61.6\% & 72.0\% & \textbf{$+10.4$} \\
        \midrule
        \textbf{Macro average} & \textbf{37.0\%} & \textbf{27.2\%} & \textbf{$-9.8$} & \textbf{67.5\%} & \textbf{71.5\%} & \textbf{$+4.0$} \\
        \bottomrule
    \end{tabularx}
\end{table*}

Table~\ref{tab:utility_preservation} reports the effect of Attention Realignment across models. 
RADAR reduces the quantized final loop rate across all models, with a  average of 27.2\% after intervention.
The largest reduction is observed on DeepSeek-R1-Distill-Qwen-14B (16.0 points), with the remaining models showing consistent improvements in the same direction.
These results support the effectiveness of attention realignment in mitigating model collapse across architectures. 
Notably, RADAR does not explicitly modify the semantic tendency toward repetition; instead, suppressing abnormal attention allocation alone is sufficient to reduce collapse, suggesting that attention redistribution plays a functional role in sustaining these failure modes. Our current evaluation focuses on targeted attacks that reliably induce uncontrolled reasoning. 
Since Attention Realignment remains effective under targeted attacks that deliberately reinforce uncontrolled generation, these results suggest that the same mechanism may also help break naturally occurring local collapse patterns. However, such spontaneous failures are relatively rare and difficult to collect at scale, preventing a systematic evaluation in the current study.

The intervention also largely preserves benign utility. Under loose answer scoring, benign accuracy decreases by only 0.8 points on DeepSeek-R1-Distill-Llama-8B and does not decrease on the other four models. This is consistent with the design of Attention Realignment, which adjusts PAS and CTS toward normal ranges rather than imposing a generic output constraint. 
A more detailed analysis of defensive performance is presented in the appendix~\ref{app:defense_comparison}.

\section{Conclusion}
\label{sec:conclusion}

Extended reasoning is difficult to control because additional computation can support productive reflection or degenerate into excessive reflection and persistent looping.
We introduced RADAR, which operationalizes this distinction as four reasoning states and tracks them from PAS and CTS trajectories with a temporal classifier.
Across five models and nine tasks, RADAR achieved an average micro-F1 of 89.8\%.
The attention analyses further identified state-dependent PAS trajectories and CTS distributions, while harmful state confidence generally strengthened before the recorded onset of repetition in naturally induced attacks.
We further used Attention Realignment as a diagnostic intervention that moves abnormal attention patterns toward those observed during normal reasoning.
Across the five evaluated models, RADAR reduced the average loop rate from 36.8\% to 27.2\%, while preserving benign accuracy on most models.
This evidence supports state-aware intervention as a way to target uncontrolled continuation without imposing the same length constraint on all reasoning.
More broadly, RADAR provides a state-level perspective on uncontrolled reasoning, moving beyond output symptoms to characterize how reasoning behavior evolves through internal attention dynamics.
By connecting state diagnosis, temporal analysis, and targeted intervention, our results suggest a path toward runtime reasoning control that distinguishes useful reflection from harmful continuation rather than treating all extended reasoning uniformly.

\section*{AI use statement}

Generative AI tools were used to produce the
\texttt{corrected\_reasoning} field in our dataset, assist with literature
retrieval, review manuscript formatting, and suggest caption and editorial
revisions. The authors reviewed all AI-assisted outputs and suggestions and
take full responsibility for the final text, data, results, and claims.

\section*{Ethics statement}

This work studies adversarial inputs that induce excessive reflection or
persistent looping in reasoning models. Such inputs can increase inference
cost and may be adapted to disrupt deployed services, so the attack procedures
and results have a dual-use dimension. We use them only in controlled
experiments to characterize failure modes and evaluate defenses. The reported
results should not be interpreted as establishing the safety of a deployment
without model- and setting-specific validation. Our experiments use existing
research benchmarks and model-generated trajectories. They involve no human
participants, private user data, or personally identifiable information.
Annotations concern reasoning behavior rather than sensitive personal
attributes. By reporting evaluation conditions, failure cases, and limitations,
we aim to support defensive research while reducing the risk of overstating the
protection offered by the evaluated intervention.

\subsection*{Reproducibility statement}

The method, online state classifier, and attention-realignment intervention are
specified in Sections~\ref{sec:method_radar}--\ref{sec:method_radar_d}, while
the evaluated models, tasks, and metrics are described in
Section~\ref{sec:exp_setup}. Appendix~\ref{app:state_annotation} provides the
reasoning-state annotation criteria and representative examples.
Appendix~\ref{app:experimental_settings} records the hardware, software,
generation parameters, and source-group configuration. The defense baselines,
fixed-candidate construction, decoding conditions, and metric definitions
are documented in Appendix~\ref{app:defense_comparison}. Finally,
Appendix~\ref{app:ablation} reports the confidence-threshold,
layer-selection, and retained-layer analyses. Tables and captions state sample
counts, inclusion rules, and denominators where cohorts differ, enabling the
reported comparisons to be reconstructed without treating conditional results
as unconditional estimates.


\bibliography{iclr2027_conference}
\bibliographystyle{iclr2027_conference}

\clearpage
\appendix
\section{Reasoning-State Annotation Criteria}
\label{app:state_annotation}

We assign the four state labels to complete trajectories used to construct the offline training set. Labels are assigned through rule-based initial classification, LLM-based verification, and final human review. Runtime predictions on partial trajectories are model outputs rather than manual annotations. The annotation protocol evaluates how a trajectory develops and terminates, not whether its final answer is correct. An incorrect response can therefore be labeled $D$ or $R$, while a correct response can be labeled $O$ if it continues reflecting after reaching a sufficient solution.

The prediction target is trajectory-level: during training, each prefix inherits its complete-trajectory label so that, at test time, the classifier estimates the eventual $D/R/O/L$ state of the ongoing trajectory from the available prefix. A prefix-level $O/L$ prediction indicates that the ongoing generation is likely to develop into an uncontrolled trajectory, rather than that excessive reflection or persistent looping has already occurred at that prefix. This early risk estimate is used to trigger attention realignment and reduce the likelihood and extent of uncontrolled output.

A \emph{reflection episode} is a contiguous span that explicitly verifies, revises, challenges, or restarts a previous intermediate conclusion. Consecutive sentences serving the same verification or reconsideration goal count as one episode. A new episode begins only after the model returns to forward problem solving or changes the object of reconsideration. We denote the number of reflection episodes in trajectory $\mathbf{y}$ by $e(\mathbf{y})$. We regard an episode as making substantive progress when it introduces a new constraint, corrects an earlier step, changes an intermediate conclusion, or supplies new evidence needed for the solution. Paraphrasing an existing statement or reaffirming it without additional support does not constitute progress.

For a generated sequence $\mathbf{y}$ of length $t$, let $\mathcal{N}_n(\mathbf{y})$ be the set of distinct $n$-grams in the sequence.
We measure the frequency share of the most frequent $n$-gram as:
\begin{equation}
    r_n(\mathbf{y})=
    \max_{g\in\mathcal{N}_n(\mathbf{y})}
    \frac{\operatorname{count}(g;\mathbf{y})}{t-n+1},
    \qquad n\in\{2,3,4\}.
\end{equation}
Let $\bar{r}^{\mathrm{normal}}_n$ denote the corresponding average over normal-request trajectories.
We define the relative repetition score as:
\begin{equation}
    \rho(\mathbf{y})=
    \max_{n\in\{2,3,4\}}
    \frac{r_n(\mathbf{y})}{\bar{r}^{\mathrm{normal}}_n}.
\end{equation}
A trajectory satisfies the lexical repetition criterion for persistent looping when $\rho(\mathbf{y})\geq 2$.

\paragraph{Direct answering ($D$).}
A trajectory is labeled $D$ when it follows a single forward solution path and reaches a definite answer without explicitly reconsidering an earlier conclusion. The quantitative boundary is $e(\mathbf{y})=0$. Ordinary elaboration, such as expanding a derivation or explaining a previously stated step, remains part of the same forward path and is not counted as reflection. The label does not require a short response and does not imply that the answer is correct. A long but continuously advancing derivation is therefore $D$, whereas any explicit verification, correction, or restart excludes the trajectory from this class.

\paragraph{Effective reflection ($R$).}
A trajectory is labeled $R$ when reconsideration remains bounded and contributes to completing the task. Operationally, it must contain between one and five reflection episodes, $1\leq e(\mathbf{y})\leq 5$, and these episodes must collectively produce substantive progress before the model reaches a definite answer. The progress requirement separates effective reflection from repeated self-confirmation: checking a calculation and correcting it qualifies, while repeatedly stating that the same calculation should be checked does not. The upper bound of five episodes provides a reproducible separation from sustained reconsideration, but episode count is interpreted together with function; cases in which the count and the progress criterion disagree are referred for adjudication.

\paragraph{Excessive reflection ($O$).}
A trajectory is labeled $O$ when reflection continues after the response has already reached a sufficient solution prefix and subsequent reconsideration contributes little or no new information. A sufficient prefix is the earliest prefix containing a complete candidate solution that could be returned as the response, irrespective of its correctness. The operational boundary is more than five reflection episodes, $e(\mathbf{y})>5$, with the later episodes repeatedly revisiting semantically similar uncertainties, derivations, or candidate answers. Unlike $R$, the defining behavior is not merely length but the saturation of useful progress. 
Unlike $L$, which repeats a relatively stable lexical pattern, $O$ forms paragraph-level reflection cycles that revisit the same semantic content while allowing variation in wording across repetitions.

\paragraph{Persistent looping ($L$).}
A trajectory is labeled $L$ when generation collapses into recurring lexical content and fails to terminate naturally before the fixed generation cap $B$. Both conditions are required: the trajectory reaches the cap and satisfies $\rho(\mathbf{y})\geq 2$. The repeated unit may be a phrase, sentence, or short token span, and minor local substitutions do not break the loop when the same pattern continues to recur. Isolated duplication, a deliberate restatement, or a repeated equation followed by normal completion is not sufficient. When a capped trajectory satisfies both the excessive-reflection and lexical-loop criteria, we assign $L$ because persistent lexical cycling is the more specific terminal behavior; $O$ is reserved for non-looping excessive reconsideration.

The resulting decision procedure first tests the two conditions for $L$, then distinguishes $D$, $R$, and $O$ using $e(\mathbf{y})$ together with the functional progress criterion. Table~\ref{tab:state_annotation_criteria} summarizes these operational boundaries.
Successful attacks are uniformly marked as \texttt{loop=True}.
\begin{table*}[t]
    \centering
    \small
    \setlength{\tabcolsep}{5pt}
    \renewcommand{\arraystretch}{1.18}
    \caption{Operational annotation criteria for the four reasoning states.}
    \label{tab:state_annotation_criteria}
    \begin{tabularx}{\textwidth}{@{}c
        >{\raggedright\arraybackslash}X
        >{\raggedright\arraybackslash}X
        >{\raggedright\arraybackslash}X@{}}
        \toprule
        \textbf{State} & \textbf{Reflection behavior} & \textbf{Output behavior} & \textbf{Quantitative cue} \\
        \midrule
        $D$ &
        No explicit reflection or backtracking. &
        Completes a single forward solution path, regardless of correctness. &
        $e(\mathbf{y})=0$. \\

        $R$ &
        Bounded verification, correction, or reconsideration that makes substantive progress. &
        Reaches a definite answer after useful reflection. &
        $1 \leq e(\mathbf{y}) \leq 5$. \\

        $O$ &
        Reconsideration continues after a sufficient solution prefix, when useful progress has saturated. &
        Remains semantically variable but adds little new information. &
        $e(\mathbf{y})>5$, without a persistent lexical loop. \\

        $L$ &
        Identical or near-identical lexical content recurs without substantive progress. &
        Fails to terminate naturally before reaching the generation cap. &
        Generation cap reached and $\rho(\mathbf{y})\geq 2$. \\
        \bottomrule
    \end{tabularx}
\end{table*}

\subsection{Illustrative Reasoning-State Examples}
\label{app:state_examples}

Table~\ref{tab:reasoning_state_examples} presents one representative trajectory from the original logs for each state. The examples complement the operational criteria above by showing how the distinctions appear in actual generations.
\begin{table*}[t]
    \centering
    \scriptsize
    \setlength{\tabcolsep}{4pt}
    \renewcommand{\arraystretch}{1.16}
    \caption{Representative examples from the original four-state trajectory logs. To control for model-specific style, all examples use DeepSeek-R1-Distill-Qwen-14B training trajectories. Quoted text is excerpted from the recorded generation; ellipses mark omitted spans. Token counts are the recorded completion lengths.}
    \label{tab:reasoning_state_examples}
    \begin{tabularx}{\textwidth}{@{}c
        >{\raggedright\arraybackslash}p{2.45cm}
        >{\raggedright\arraybackslash}X
        >{\raggedright\arraybackslash}p{3.15cm}@{}}
        \toprule
        \textbf{State} & \textbf{Source record} & \textbf{Prompt and trajectory excerpt} & \textbf{Label evidence} \\
        \midrule
        \rowcolor{gray!12}
        $D$ &
        GSM8K &
        \textbf{Prompt:} Janet's ducks lay 16 eggs per day; after eating three and using four for muffins, how much does she earn by selling the rest for \$2 each? \textbf{Excerpt:} ``Total eggs used~$=3+4=7$~eggs. \ldots{} Eggs for sale~$=16-7=9$~eggs. \ldots{} Daily earnings~$=9\times\$2=\$18$. \ldots{} Final Answer: $18$.'' &
        One forward calculation reaches a definite answer and terminates normally after 488 tokens, without explicit verification or backtracking. \\

        $R$ &
        GPQA &
        \textbf{Prompt:} In a universe with isolated magnetic poles, which Maxwell equation changes? \textbf{Excerpt:} ``The equation is $\nabla\!\cdot\!B=0$, which \ldots{} states that there are no isolated magnetic poles. \ldots{} Wait, but let me double-check. \ldots{} The main change would be in the divergence of $B$ \ldots{} So the correct answer should be option C.'' &
        A bounded double-check examines a possible alternative, confirms the initial physical argument, and ends with a definite answer after 722 tokens. \\

        \rowcolor{gray!12}
        $O$ &
        Recur &
        \textbf{Prompt:} Infer a nurses' station's starting stock from three days of bandage use and orders, while explaining why 19 is incorrect. \textbf{Excerpt:} ``At the end, $X+59=78 \Rightarrow X=19$. But starting with 19, using 38 is impossible. \ldots{} Therefore, the problem is flawed. \ldots{} However, \ldots{} perhaps the correct answer is different. Maybe I made a mistake in the equations. Let me try to set up the equations again, carefully.'' &
        The trajectory repeatedly revisits delivery timing and the same equations after useful progress has saturated. It remains semantically variable, produces no final answer, and reaches the 16,384-token cap without collapsing to one fixed lexical span. \\

        $L$ &
        LoopLLM &
        \textbf{Prompt:} Remove all vowels from a sentence containing an adversarial suffix. \textbf{Excerpt:} ``Wait, but in the original, it's `\texttt{* * af Zh * * \ldots{}}', so after processing, it's `\texttt{* * f Zh * * \ldots{}}'. \ldots{} So the final processed sentence is: `\texttt{Thqckbrwnfxjmpsrvthlzydg. * * f Zh * * \ldots{}}'.'' The same clauses and symbol span then recur. &
        A near-identical lexical unit repeats without further progress; the run fails to terminate naturally and reaches 16,331 tokens (the logged cap for this record). \\
        \bottomrule
    \end{tabularx}
\end{table*}

\section{Detailed Four-State Detection Results}
\label{app:four_state_detailed}

This section reports the model--dataset results underlying
Figure~\ref{fig:four_state_accuracy}. For each trajectory, RADAR predicts the
state with the largest posterior probability when that probability exceeds
0.5; otherwise, the trajectory is treated as unclassified. Attack trajectories
without an observed loop are assigned their corresponding benign $D$ or $R$
label. In Table~\ref{tab:four_state_dataset_accuracy}, each cell has the form
$a\%\,(S)$, where $a$ is classification accuracy and $S$ is the state with the
largest mean confidence within that model--dataset group. The failed-attack
rows are reported separately rather than being merged into the successful
attack results.

\begin{table*}[t]
    \centering
    \small
    \caption{Four-state classification accuracy and confidence state by model, dataset, and attack outcome. N/A indicates that the model--dataset outcome group contains no samples.}
    \label{tab:four_state_dataset_accuracy}
    \resizebox{\textwidth}{!}{%
    \rowcolors{2}{gray!10}{white}
    \begin{tabular}{lccccc}
        \toprule
        \rowcolor{gray!25}
        Dataset & DS-Llama-8B & DS-Qwen-14B & QwQ-32B & Qwen3.6-27B & GLM-4.7-Flash \\
        \midrule
        \rowcolor{gray!18}
        \multicolumn{6}{l}{\textit{Normal trajectories}} \\
        GSM8K & 100.0\% (D) & 100.0\% (D) & 100.0\% (D) & 100.0\% (D) & 96.0\% (D) \\
        MMLU-Geor & 96.0\% (D) & 96.0\% (D) & 96.0\% (D) & 100.0\% (D) & 96.0\% (D) \\
        GPQA & 100.0\% (R) & 100.0\% (R) & 100.0\% (R) & 100.0\% (R) & 100.0\% (R) \\
        MMLU-Econometrics & 48.0\% (D) & 68.0\% (R) & 50.0\% (R) & 68.0\% (R) & 52.0\% (R) \\
        MMLU-World-History & 100.0\% (R) & 100.0\% (R) & 100.0\% (R) & 100.0\% (R) & 100.0\% (R) \\
        \midrule
        \rowcolor{gray!18}
        \multicolumn{6}{l}{\textit{Successful attacks}} \\
        Recur & 100.0\% (O) & 100.0\% (O) & 50.0\% (O) & 100.0\% (O) & N/A \\
        LoopLLM & 100.0\% (L) & 80.0\% (L) & N/A & N/A & 100.0\% (L) \\
        Joint & 100.0\% (L) & 100.0\% (L) & 100.0\% (L) & 96.0\% (L) & 100.0\% (L) \\
        MiP & 100.0\% (L) & 100.0\% (L) & 100.0\% (L) & 100.0\% (L) & 100.0\% (L) \\
        \midrule
        \rowcolor{gray!18}
        \multicolumn{6}{l}{\textit{Failed attacks, evaluated with benign labels}} \\
        Recur & N/A & 18.2\% (O) & 0.0\% (O) & 0.0\% (O) & 100.0\% (O) \\
        LoopLLM & 33.3\% (D) & 85.0\% (D) & 100.0\% (D) & 100.0\% (D) & N/A \\
        \bottomrule
    \end{tabular}%
    }
\end{table*}

\subsection{Micro-F1}

Let $TP_c$, $FP_c$, and $FN_c$ denote the class-specific counts for
$c\in\mathcal{S}=\{D,R,O,L\}$. We aggregate these counts before computing
precision and recall:
\begin{equation}
    P_{\mathrm{micro}}
    =\frac{\sum_{c\in\mathcal{S}}TP_c}
    {\sum_{c\in\mathcal{S}}(TP_c+FP_c)},
    \qquad
    R_{\mathrm{micro}}
    =\frac{\sum_{c\in\mathcal{S}}TP_c}
    {\sum_{c\in\mathcal{S}}(TP_c+FN_c)}.
\end{equation}
The reported score is
\begin{equation}
    F1_{\mathrm{micro}}
    =\frac{2P_{\mathrm{micro}}R_{\mathrm{micro}}}
    {P_{\mathrm{micro}}+R_{\mathrm{micro}}}.
\end{equation}
For this single label evaluation, micro-F1 is equal to overall
classification accuracy. Table~\ref{tab:four_state_micro_f1} reports the
model-level results.
Under the standard classification interpretation of this metric,
micro-F1 ranges from 0 to 1, with larger values indicating more correct
predictions and 1 denoting perfect classification~\citep{sokolova2009systematic}.
Because it coincides with accuracy in our setting, the average score of 89.8\%
means that RADAR assigns the correct reasoning state to nearly nine out of ten
trajectories.

\begin{table}[t]
    \centering
    \small
    \caption{Model-level micro-F1 for four-state detection.}
    \label{tab:four_state_micro_f1}
    \rowcolors{2}{gray!10}{white}
    \begin{tabular}{lr}
        \toprule
        \rowcolor{gray!25}
        Model & Micro-F1 \\
        \midrule
        DeepSeek-R1-Distill-Llama-8B & 92.0\% \\
        DeepSeek-R1-Distill-Qwen-14B & 89.9\% \\
        QwQ-32B & 84.3\% \\
        Qwen3.6-27B & 88.8\% \\
        GLM-4.7-Flash & 93.8\% \\
        \midrule
        \rowcolor{gray!18}
        Average & 89.8\% \\
        \bottomrule
    \end{tabular}
\end{table}

\subsection{GLM-4.7-Flash on Recur}

Recur~\citep{wang2026recur} does not produce successful looping outcomes on GLM-4.7-Flash under our recorded output-level criterion: none of the test trajectories is marked as a successful loop, and their evaluation labels are therefore restored to benign states.
Nevertheless, RADAR assigns all trajectories to $O$, with a mean $O$ confidence of 86.0\%. This discrepancy should not be interpreted simply as a classification failure. For GLM-4.7-Flash, the $O$ class is learned from Recur trajectories during training, so the classifier is explicitly exposed to the attention dynamics associated with Recur-induced excessive reflection. The test trajectories are drawn from the same attack mechanism and consequently match this learned $O$ pattern closely, even though the attack does not ultimately satisfy the output-level looping criterion.
In other words, the classifier captures an internal excessive-reflection signature that can remain present without developing into an observable loop.

Figure~\ref{fig:glm_recur_nonloop_example} illustrates this boundary with a trajectory that terminates normally but continues with semantically unproductive reconsideration after reaching the correct result.

\begin{figure*}[t]
    \centering
    \fbox{%
    \begin{minipage}{0.95\textwidth}
    \small
    \textbf{Task.}
    A bag contains 25 chocolate bars and 80 candied apples. Each chocolate bar weighs 40~g and twice as much as an apple. The prompt asks why the answer is 1937~g.

    \par\smallskip
    \textbf{Correct resolution reached early.}
    The model derives an apple weight of 20~g and the correct total
    $25\times40+80\times20=2600$~g.

    \par\smallskip
    \textbf{Continued reasoning excerpt.}
    \begin{quote}
    \emph{Maybe the user copied the numbers wrong. Maybe there are fewer apples? Maybe there are fewer chocolate bars? \ldots{} Maybe the bag itself weighs something? No, that adds weight. Maybe the ``25'' is actually ``75''? Maybe the ``80'' is actually ``40''? \ldots{} Okay, the math is solid. \ldots{} Wait, what if the user meant ``Each chocolate bar weighs 40g MORE than each candied apple''?}
    \end{quote}

    \textbf{Observed pattern.}
    The trajectory does not repeat a fixed lexical unit, yet it keeps proposing unsupported variants and reconfirming an already settled calculation. It is therefore a non-looping instance of semantically unproductive over-reflection rather than a normal concise solution.
    \end{minipage}}
    \caption{Illustrative GLM-4.7-Flash generation on Recur. The generation stops naturally after 2,583 completion tokens and is recorded as \texttt{loop=false}. The excerpt is shortened to preserve the characteristic pattern without reproducing the full trajectory.}
    \label{fig:glm_recur_nonloop_example}
\end{figure*}

\clearpage

\section{Generation Length as an Auxiliary Classification Signal}
\label{app:length_signal}

The Temporal State Inference Classifier includes the current generation progress, $f_{i,3}=\log k_i$, which raises a potential shortcut concern: the classifier might distinguish benign and uncontrolled reasoning primarily from elapsed output length rather than from attention dynamics. We examine this possibility using the completion lengths of benign trajectories, the first-crossing times in Table~\ref{tab:detection_timeliness}, and the temporal patterns in Figures~\ref{fig:detection_accuracy_by_length} and~\ref{fig:pas_slope_distribution}. Together, these results are inconsistent with a classifier whose decisions are determined by absolute generation length alone.

\begin{table*}[t]
    \centering
    \small
    \setlength{\tabcolsep}{7pt}
    \renewcommand{\arraystretch}{1.15}
    \caption{Mean recorded completion length in tokens for the two benign reasoning states. $D$ pools GSM8K and MMLU-Geor, $R$ pools GPQA, MMLU-Econometrics, and MMLU-World-History, and $D/R$ is the sample-weighted mean over both states.}
    \label{tab:normal_generation_lengths}
    \begin{tabular}{l|ccc}
        \toprule
        \textbf{Model} & \textbf{$D$} & \textbf{$R$} & \textbf{$D/R$} \\
        \midrule
        DeepSeek-R1-Distill-Llama-8B & 638.58 & 1,836.56 & 1,357.37 \\
        DeepSeek-R1-Distill-Qwen-14B & 668.54 & 1,528.44 & 1,184.48 \\
        QwQ-32B & 1,384.90 & 1,845.13 & 1,661.04 \\
        GLM-4.7-Flash & 987.12 & 1,991.77 & 1,589.91 \\
        Qwen3.6-27B & 1,246.24 & 2,702.08 & 2,119.74 \\
        \bottomrule
    \end{tabular}
\end{table*}

Normal reasoning already spans long outputs. Table~\ref{tab:normal_generation_lengths} shows that the mean completion length of $R$ is between 1,528.44 and 2,702.08 tokens across the five models. Even after pooling $D$ and $R$, every model has a benign mean above 1,184 tokens. Long generation is therefore not specific to uncontrolled reasoning: a prefix can occur well into a normal solution while the model is still making useful progress. This is also consistent with our functional state definitions, under which $D$ denotes a continuously advancing solution path rather than a short response, and $R$ explicitly permits extended but productive reconsideration.

The detection timeline provides a second distinction between elapsed length and harmful-state evidence. In Table~\ref{tab:detection_timeliness}, the mean first position at which $P(O)+P(L)\geq 0.5$ ranges from 60.6 to 997.0 tokens across the four attack groups. Every value is below the smallest model-level benign $D/R$ mean in Table~\ref{tab:normal_generation_lengths}; Recur, Joint, and MiP cross the threshold by 450.4 tokens, and Recur and Joint do so before 100 tokens. For the naturally induced attacks, these moderate-confidence crossings also precede the recorded onset of repetition. 
These results show that harmful-state evidence can become apparent while generation is still within the normal length range, indicating that detection is not simply driven by elapsed output length.

Finally, the classifier uses length to contextualize attention evolution rather than as a substitute for it. Figure~\ref{fig:detection_accuracy_by_length} shows different confidence trajectories at comparable stages of generation: $O$ is recognizable early, whereas $L$ initially overlaps with $D$ and becomes distinguishable only as its trajectory develops. Figure~\ref{fig:pas_slope_distribution} provides the corresponding attention evidence. PAS changes with generation progress because the generated context expands, but its direction and rate of change remain state dependent. The progress feature $\log k_i$ therefore supplies a temporal coordinate for interpreting the PAS trend and mean CTS, while PAS and CTS characterize how attention is allocated at that stage. These observations support the narrower conclusion that generation length is an informative auxiliary feature but is not sufficient to determine the four-state prediction.

\section{Cross-Model CTS Distributions}
\label{app:cts_cross_model}

\begin{figure*}[t]
    \centering
    \begin{minipage}[t]{0.49\textwidth}
        \centering
        \includegraphics[width=\linewidth]{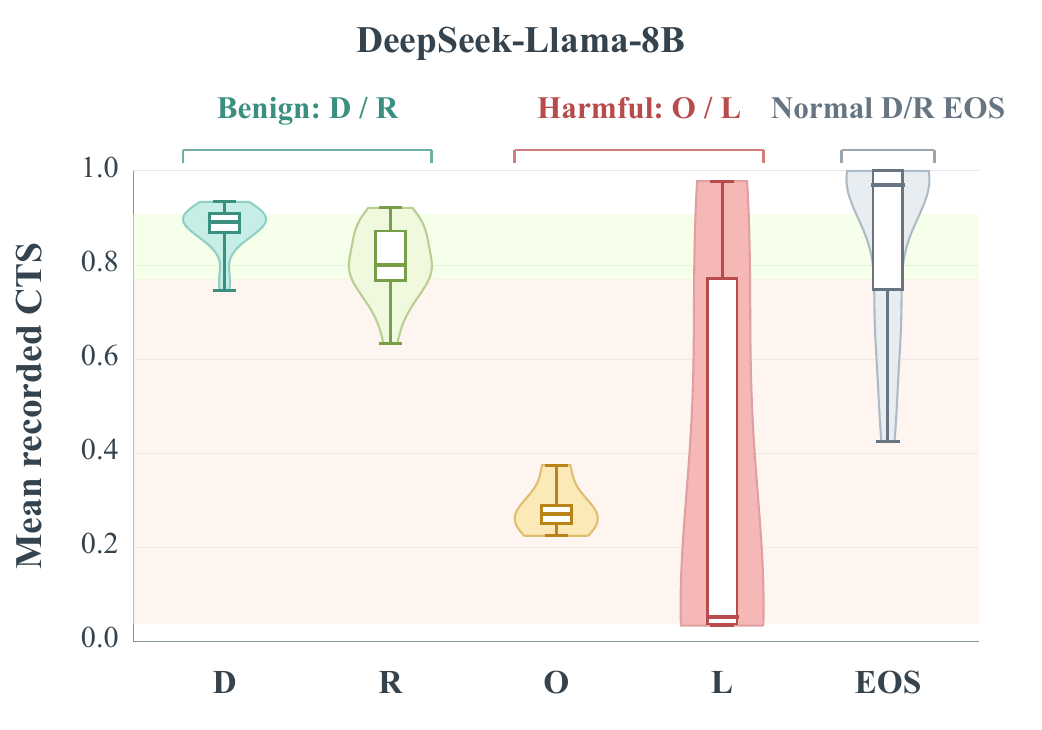}
        \par\smallskip
        \textbf{(a) DeepSeek-R1-Distill-Llama-8B}
    \end{minipage}\hfill
    \begin{minipage}[t]{0.49\textwidth}
        \centering
        \includegraphics[width=\linewidth]{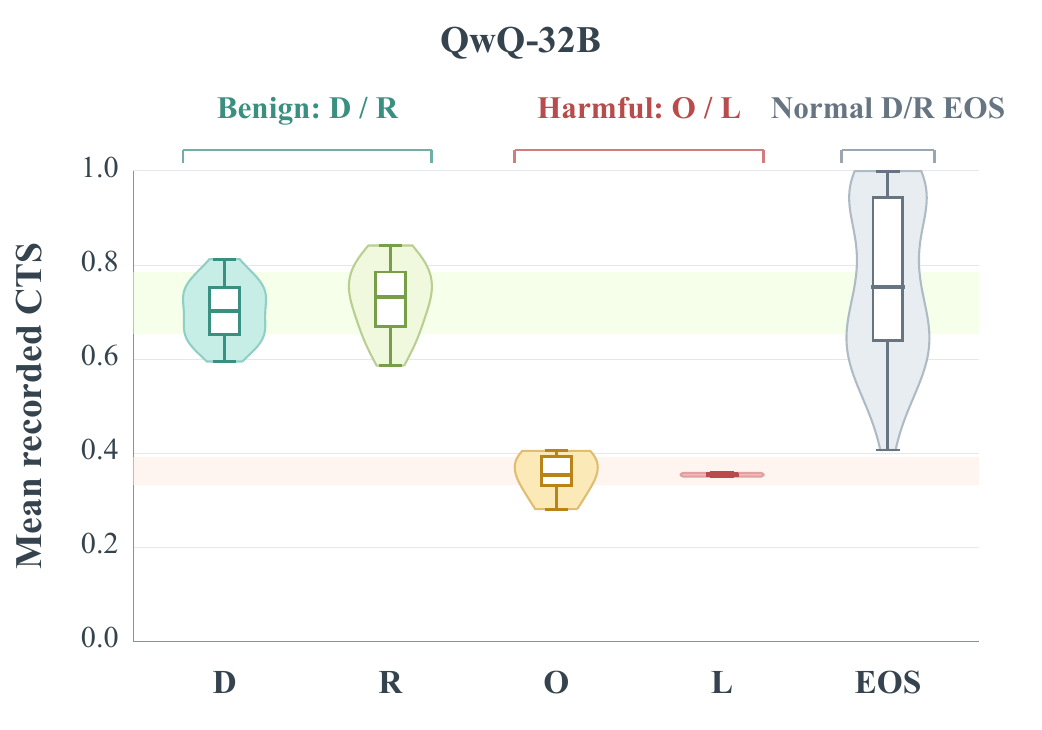}
        \par\smallskip
        \textbf{(b) QwQ-32B}
    \end{minipage}
    \caption{Supplementary CTS distributions for DeepSeek-R1-Distill-Llama-8B and QwQ-32B. The first four violins show per-trajectory mean recorded CTS for D/R/O/L; the EOS violin shows the final recorded CTS for normal D/R trajectories. Boxes denote interquartile ranges, center lines denote medians, and whiskers denote observed extrema.}
    \label{fig:cts_cross_model_appendix_a}
\end{figure*}

\begin{figure*}[t]
    \centering
    \begin{minipage}[t]{0.49\textwidth}
        \centering
        \includegraphics[width=\linewidth]{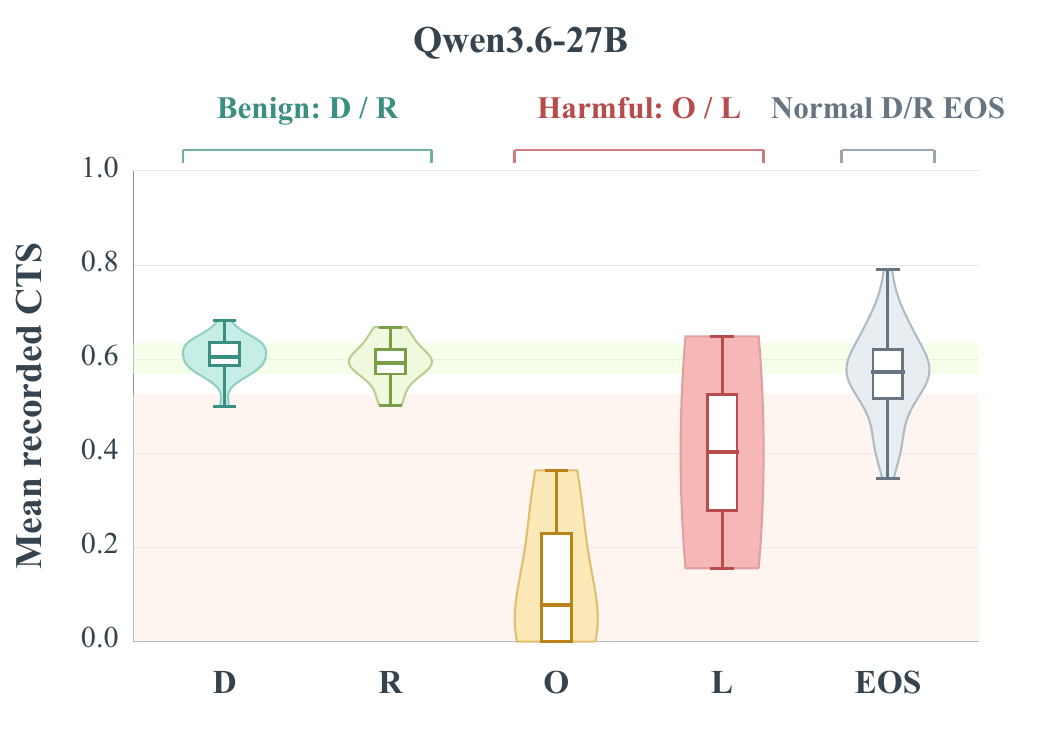}
        \par\smallskip
        \textbf{(c) Qwen3.6-27B}
    \end{minipage}\hfill
    \begin{minipage}[t]{0.49\textwidth}
        \centering
        \includegraphics[width=\linewidth]{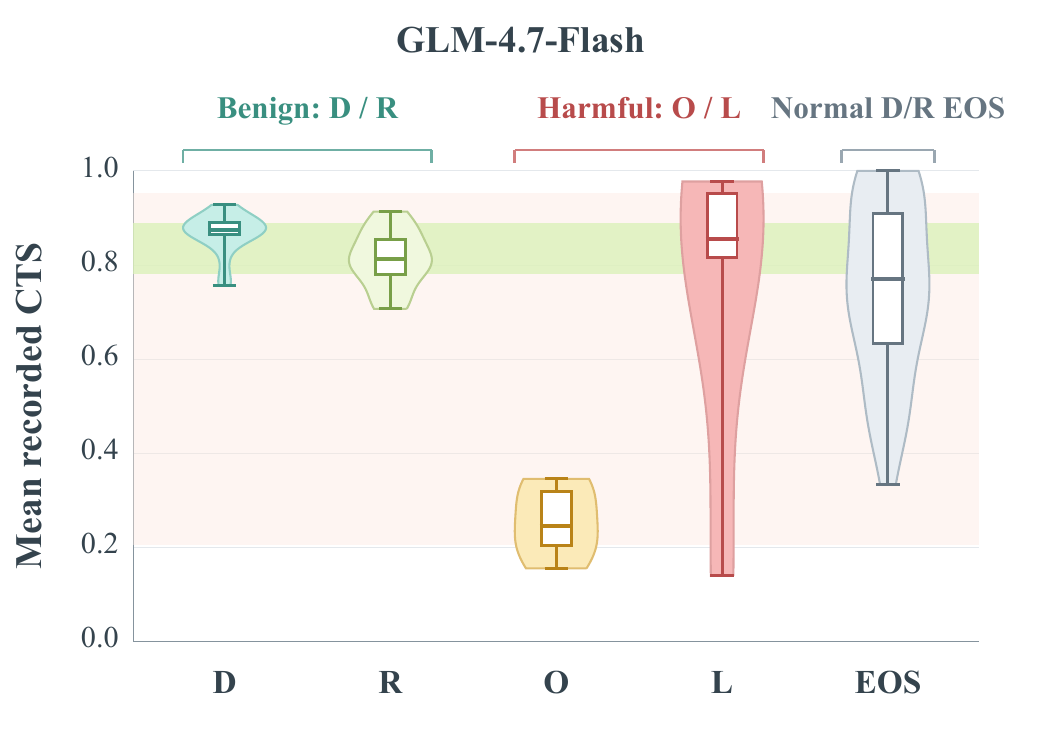}
        \par\smallskip
        \textbf{(d) GLM-4.7-Flash}
    \end{minipage}
    \caption{Supplementary CTS distributions for Qwen3.6-27B and GLM-4.7-Flash, using the same aggregation, EOS control, and visual conventions as Figure~\ref{fig:cts_cross_model_appendix_a}.}
    \label{fig:cts_cross_model_appendix_b}
\end{figure*}

Figures~\ref{fig:cts_cross_model_appendix_a} and \ref{fig:cts_cross_model_appendix_b} extend the CTS analysis in Section~\ref{sec:exp_mechanism_transfer} to the four remaining model families.
All panels use the same state-specific training sources as the main analysis: GSM8k for $D$, GPQA for $R$, Recur for $O$, and LoopLLM for $L$.
The state violins summarize the per-trajectory mean recorded CTS, whereas the EOS violin contains the final recorded CTS from the  normal $D/R$ trajectories.

Across models, the dominant pattern is that normal $D/R$ requests and collapsed $O/L$ trajectories are separable through their cross-context token allocation.
In particular, the mean CTS for $O$ remains below the corresponding $D$ and $R$ means in all five models.
The separation is especially consistent for the Qwen family: the main-text DeepSeek-Qwen-14B result and the supplementary QwQ-32B and Qwen3.6-27B panels all place the harmful-state distributions below the benign pair.
The small number of $L$ samples for QwQ-32B and Qwen3.6-27B, however, prevents a strong variance claim for those two panels.

The EOS controls further show that termination alone does not induce the low-CTS pattern associated with collapse.
Because EOS contributes one terminal measurement per trajectory rather than an average over the full trajectory, its distribution is wider than the D/R trajectory-mean distributions.
Even so, the EOS interquartile interval overlaps the benign D/R range in every model, and its mean remains substantially closer to the benign states than to $O$.
Normal requests therefore retain broadly aligned prompt--generation attention over shared token identities even at their final recorded step.

GLM-4.7-Flash exposes an informative boundary to this general separation.
Its $O$ trajectories retain the usual low-CTS signature, but its $L$ mean CTS is 0.75, close to the benign means of 0.87 for $D$ and 0.81 for $R$.
Persistent looping in GLM can therefore preserve a comparatively balanced relative distribution over token identities shared by the prompt and generation.
Figure~\ref{fig:pas_slope_distribution} provides the complementary signal: GLM's length-normalized PAS for $L$ rises from 0.93 at the start to 44.25 at 4,000 generated tokens, while $O$ remains near 1.10.
Taken together, high CTS and sharply increasing PAS suggest that the loop remains strongly anchored to prompt-side occurrences of token identities that are repeatedly reused in the output.
This joint interpretation explains how GLM can exhibit apparently balanced cross-context token distributions and still collapse: the failure is not an input--output identity mismatch, but excessive absolute attention to a prompt-anchored repetitive token set.

\paragraph{Empty-overlap boundary case.}
Because the CTS expression in Section~\ref{sec:method_cts} sums over token identities shared by the prompt and generated context, an empty intersection makes the sum vanish and yields $M_{\mathcal C}(s)=1$.
We did not observe this condition in any trajectory used in our experiments.
It requires an extreme output prefix with no token identity shared with the prompt; this is particularly unlikely for the long, semantically rich trajectories produced under reasoning collapse, because a longer context provides more opportunities for shared identities.
If the condition occurs in a benign trajectory, the resulting value is also consistent with the high-CTS regime observed for normal $D/R$ examples.
Accordingly, this boundary case does not affect the reported empirical results.

\section{Supplementary Experimental Settings}
\label{app:experimental_settings}

\subsection{Reasoning Trajectory Generation}
\label{app:trajectory_generation}

\paragraph{Hardware and execution backend.}
Reasoning trajectories are generated on a single NVIDIA A100-SXM4 GPU with 80\,GB of memory, running Linux 6.11.0-25-generic. The vLLM GPU memory utilization parameter is set to 0.9 (\texttt{--gpu-memory-utilization 0.9}). With \texttt{--backend auto}, the generation pipeline checks the model's \texttt{architectures} field in \texttt{config.json} against the vLLM model registry to select either vLLM or Hugging Face Transformers as the execution backend.

\paragraph{Software environment.}
Table~\ref{tab:trajectory_software} lists the software versions used for trajectory generation.
\begin{table}[htbp]
    \centering
    \caption{Software environment for reasoning trajectory generation.}
    \label{tab:trajectory_software}
    \begin{tabular}{ll}
        \toprule
        Component & Version \\
        \midrule
        Python & 3.12.9 \\
        PyTorch & 2.7.0+cu126 \\
        CUDA & 12.6 \\
        cuDNN & 9.5.1 \\
        Transformers & 4.52.3 \\
        vLLM & 0.9.0 \\
        Tokenizers & 0.21.1 \\
        Accelerate & 1.9.0 \\
        NumPy & 1.26.4 \\
        \bottomrule
    \end{tabular}
\end{table}

\paragraph{Generation parameters and source groups.}
All four source groups use a sampling temperature of 0.5, a maximum generation budget of 16k tokens (\texttt{max\_tokens}), and a base random seed of 0 (\texttt{base\_seed}). Table~\ref{tab:trajectory_source_groups} records the configuration identifiers and their corresponding data sources. These identifiers describe trajectory collection groups; final reasoning-state labels follow the annotation criteria in Appendix~\ref{app:state_annotation}.
\begin{table}[htbp]
    \centering
    \small
    \caption{Trajectory collection groups. All groups share temperature 0.5, a 16k-token generation cap, and base seed 0. Configuration identifiers are retained for reproducibility.}
    \label{tab:trajectory_source_groups}
    \begin{tabularx}{\textwidth}{lX}
        \toprule
        Configuration identifier & Data sources \\
        \midrule
        \texttt{concise\_reasoning} & GSM8K, MMLU-Geor \\
        \texttt{productive\_reasoning} & GPQA, Econometrics, World\_History \\
        \texttt{repetitive\_reasoning} & RECUR \\
        \texttt{repetitive\_string} & LoopLLM \\
        \texttt{repetitive\_string} & Joint \\
        \texttt{repetitive\_string} & Missing Premise (MiP) \\
        \bottomrule
    \end{tabularx}
\end{table}
\section{Cross-Defense Comparison}
\label{app:defense_comparison}

This appendix compares RADAR with output-oriented defenses on
DeepSeek-R1-Distill-Llama-8B. The comparison supplements the cross-model
intervention results in Section~\ref{sec:exp_defense} by isolating how different
defense mechanisms suppress uncontrolled generation on the same model and
fixed attack cohorts.

\subsection{Compared Defenses and Evaluation Protocol}

\paragraph{Defense baselines.}
We compare four generation conditions. \emph{Undefended} uses the original
decoding configuration. \emph{RADAR} activates the attention-realignment
intervention from Section~\ref{sec:method_radar_d} when the online state
classifier identifies harmful reasoning. \emph{AUSteer}~\citep{austeer2026}
is an activation-steering baseline configured with $k=16$ selected atomic
units and steering strength $\alpha=10$. \emph{n-gram} is an output-level
decoding constraint that prevents a generated trigram from
recurring~\citep{paulus2018deep,guan-huang-2023-mitigating}. The
latter two baselines act directly on activation-derived output scores or the
generated token sequence, whereas RADAR first exposes a reasoning-state
signal and uses the diagnosed attention pattern to determine when and where to
intervene.

\subsection{Attack Suppression}
\begin{table*}[t]
    \centering
    \small
    \setlength{\tabcolsep}{4pt}
    \renewcommand{\arraystretch}{1.15}
    \caption{Cross-defense loop rates on DeepSeek-R1-Distill-Llama-8B. $\Delta$ is the percentage-point change from Undefended; lower values are better.}
    \label{tab:attack_suppression}
    \begin{tabular}{l c cc cc cc}
        \toprule
        & \textbf{Undefended}
        & \multicolumn{2}{c}{\textbf{RADAR}}
        & \multicolumn{2}{c}{\textbf{AUSteer}}
        & \multicolumn{2}{c}{\textbf{n-gram}} \\
        \cmidrule(lr){2-2}\cmidrule(lr){3-4}\cmidrule(lr){5-6}\cmidrule(l){7-8}
        \textbf{Attack} & \textbf{Rate} & \textbf{Rate} & \textbf{$\Delta$ (pp)} & \textbf{Rate} & \textbf{$\Delta$ (pp)} & \textbf{Rate} & \textbf{$\Delta$ (pp)} \\
        \midrule
        \rowcolor[HTML]{F3F4F6} Recur & 56.0\% & 44.0\% & $-12.0$ & 48.0\% & $-8.0$ & 0.0\% & $-56.0$ \\
        LoopLLM & 56.0\% & 28.0\% & $-28.0$ & 20.0\% & $-36.0$ & 0.0\% & $-56.0$ \\
        \rowcolor[HTML]{F3F4F6} MiP & 56.0\% & 48.0\% & $-8.0$ & 44.0\% & $-12.0$ & 0.0\% & $-56.0$ \\
        \midrule
        \textbf{Macro average} & \textbf{56.0\%} & \textbf{40.0\%} & \textbf{$-16.0$} & \textbf{37.3\%} & \textbf{$-18.7$} & \textbf{0.0\%} & \textbf{$-56.0$} \\
        \bottomrule
    \end{tabular}
\end{table*}

Table~\ref{tab:attack_suppression} reports final loop rate using the same
percentage-based metric as the main defense results. The undefended macro
average is 56.0\%. RADAR lowers it to 40.0\%, a reduction of 16.0 percentage
points, while AUSteer reaches 37.3\% ($-18.7$ points) and n-gram reaches 0.0\%
($-56.0$ points). The attack-specific results show that RADAR has its largest
effect on LoopLLM, where the final loop rate falls from 56.0\% to 28.0\%.
RADAR is therefore not the strongest method under this output-level suppression
metric. Its value in this comparison is that the intervention is tied to an explicit diagnosis of
the reasoning state and the associated attention dynamics, rather than to a
generic constraint on the output distribution or surface repetition. In other
words, methods that directly reshape output scores or prohibit repeated token
patterns achieve more aggressive suppression, whereas RADAR applies a more
conservative, state-conditioned correction. 

\subsection{Benign-Task Utility}
\begin{table}[t]
    \centering
    \small
    \setlength{\tabcolsep}{7pt}
    \renewcommand{\arraystretch}{1.15}
    \caption{Cross-defense benign-task utility on DeepSeek-R1-Distill-Llama-8B. Accuracy is measured on the same five normal-task subsets used in the main utility comparison, excluding SimpleQA. $\Delta$ is the percentage-point change from Undefended.}
    \label{tab:cross_defense_benign_utility}
    \begin{tabular}{lcc}
        \toprule
        \textbf{Method} & \textbf{Benign accuracy $\uparrow$} & \textbf{$\Delta$ (pp)} \\
        \midrule
        \rowcolor[HTML]{F3F4F6} Undefended & 66.4\% & -- \\
        RADAR & 65.6\% & $-0.8$ \\
        \rowcolor[HTML]{F3F4F6} AUSteer & 66.4\% & $0.0$ \\
        n-gram & 58.4\% & $-8.0$ \\
        \bottomrule
    \end{tabular}
\end{table}

Table~\ref{tab:cross_defense_benign_utility} complements the attack-only
comparison with answer accuracy on normal tasks under the same four generation
conditions. The undefended model attains 66.4\% accuracy.
RADAR changes this result to 65.6\%, a decrease of 0.8 percentage points, while
AUSteer remains at 66.4\%. In contrast, n-gram blocking lowers benign accuracy
to 58.4\%, an 8.0-point decrease, while also producing the lowest final loop
rate in Table~\ref{tab:attack_suppression}. This contrast shows that an aggressive external decoding constraint can
obtain stronger attack suppression by changing the model's behavior on normal
problems.
Overall, external interventions can alter normal-task performance, and the
magnitude of this effect depends on the intervention; defense effectiveness
should therefore be assessed jointly with benign utility rather than from
attack suppression alone.

\section{Attack-Type Suppression across Models}
\label{app:attack_type_suppression}

We further disaggregate RADAR by model and natural attack type, focusing on Recur, LoopLLM, and MiP because together they cover the main forms of naturally induced uncontrolled reasoning and model collapse considered in our evaluation.
\begin{table*}[t]
    \centering
    \small
    \setlength{\tabcolsep}{3pt}
    \renewcommand{\arraystretch}{1.10}
    \caption{Attack-type suppression by RADAR across models. Overall reports the average across the five models. Joint is excluded because it does not represent a naturally induced attack.}
    \label{tab:attack_type_suppression_cross_model}
    \begin{tabularx}{\textwidth}{>{\raggedright\arraybackslash}Xl|ccccc|c}
        \toprule
        \textbf{Attack} & \textbf{Setting} & \textbf{Llama-8B} & \textbf{Qwen-14B} & \textbf{QwQ-32B} & \textbf{Qwen3.6} & \textbf{GLM-4.7} & \textbf{Overall} \\
        \midrule
        Recur & Undefended Loop & 56.0\% & 12.0\% & 8.0\% & 4.0\% & 0.0\% & 16.0\% \\
        & RADAR Loop & 44.0\% & 8.0\% & 4.0\% & 4.0\% & 0.0\% & 12.0\% \\
        \midrule
        LoopLLM & Undefended Loop & 56.0\% & 36.0\% & 0.0\% & 0.0\% & 80.0\% & 34.4\% \\
        & RADAR Loop & 28.0\% & 0.0\% & 0.0\% & 0.0\% & 36.0\% & 12.8\% \\
        \midrule
        MiP & Undefended Loop & 56.0\% & 64.0\% & 80.0\% & 60.0\% & 44.0\% & 60.8\% \\
        & RADAR Loop & 48.0\% & 56.0\% & 68.0\% & 48.0\% & 44.0\% & 52.8\% \\
        \bottomrule
    \end{tabularx}
\end{table*}

Table~\ref{tab:attack_type_suppression_cross_model} shows that RADAR suppresses
all three attack categories after pooling the five models. The Loop rate falls
from 16.0\% to 12.0\% for Recur, from 34.4\% to 12.8\% for LoopLLM, and from
60.8\% to 52.8\% for MiP.
Overall, the aggregate evidence supports effectiveness on uncontrolled reasoning.

\section{Ablation Studies}
\label{app:ablation}

\subsection{Detection Confidence Threshold}

\begin{figure*}[t]
    \centering
    \includegraphics[
        width=\textwidth,
        height=0.48\textheight,
        keepaspectratio
    ]{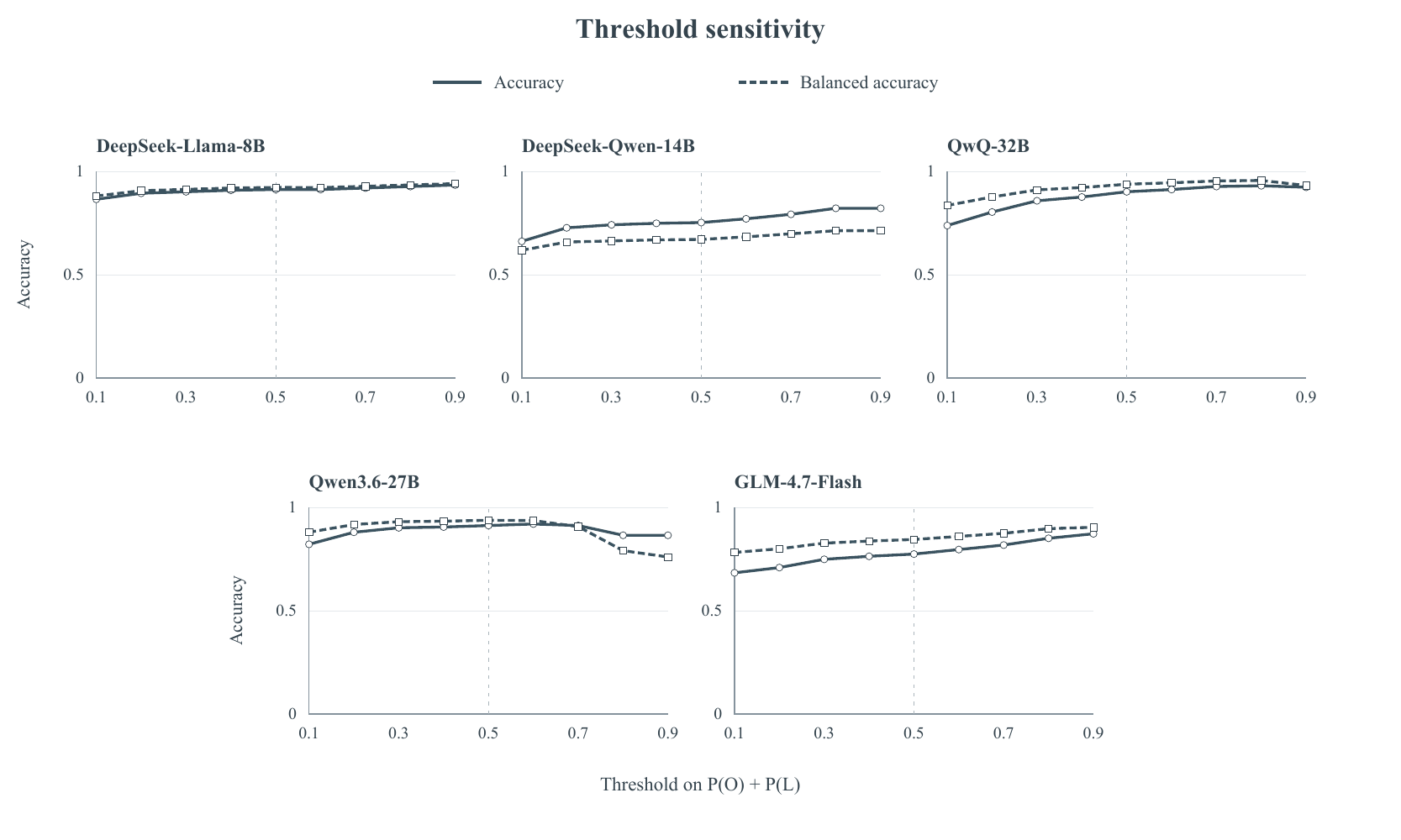}
    \caption{
    Binary accuracy and balanced accuracy versus the confidence threshold
    at each trajectory's terminal power-of-two checkpoint.
    A prediction is harmful when $P(O)+P(L)>\tau$.
    O/L are positive classes and D/R are negative classes.
    Balanced accuracy averages positive and negative recall.
    }
    \label{fig:threshold_accuracy}
\end{figure*}

\begin{table*}[t]
\centering
\small
\setlength{\tabcolsep}{4pt}
\renewcommand{\arraystretch}{1.15}
\caption{Binary detection accuracy at different classifier confidence thresholds. A trajectory is predicted as harmful when $P(O)+P(L)>\tau$.}
\label{tab:classifier_threshold_accuracy}
\begin{tabular}{cccccc}
\toprule
\textbf{$\tau$} & \textbf{Llama-8B} & \textbf{Qwen-14B} & \textbf{QwQ-32B} & \textbf{Qwen3.6-27B} & \textbf{GLM-4.7} \\
\midrule
\rowcolor[HTML]{F3F4F6} 0.1 & 86.55\% & 66.18\% & 73.82\% & 82.18\% & 68.36\% \\
0.2 & 89.45\% & 72.73\% & 80.36\% & 88.00\% & 70.91\% \\
\rowcolor[HTML]{F3F4F6} 0.3 & 90.18\% & 74.18\% & 85.82\% & 90.18\% & 74.91\% \\
0.4 & 90.91\% & 74.91\% & 87.64\% & 90.55\% & 76.36\% \\
\rowcolor[HTML]{F3F4F6} 0.5 & 91.27\% & 75.27\% & 90.18\% & 91.27\% & 77.45\% \\
0.6 & 91.27\% & 77.09\% & 91.27\% & \textbf{92.00\%} & 79.64\% \\
\rowcolor[HTML]{F3F4F6} 0.7 & 92.00\% & 79.27\% & 92.73\% & 91.27\% & 81.82\% \\
0.8 & 92.73\% & \textbf{82.18\%} & \textbf{93.09\%} & 86.55\% & 85.09\% \\
\rowcolor[HTML]{F3F4F6} 0.9 & \textbf{93.45\%} & \textbf{82.18\%} & 92.36\% & 86.55\% & \textbf{87.27\%} \\
\bottomrule
\end{tabular}
\end{table*}

Figure~\ref{fig:threshold_accuracy} shows how ordinary and balanced accuracy vary with the terminal detection threshold.
For most models, both metrics improve as the threshold increases, although the optimal operating point remains model dependent: QwQ-32B peaks near $\tau=0.8$, while Qwen3.6-27B performs best at a moderate threshold and degrades at higher values. 
This confirms that stricter thresholds do not universally improve detection and that class-wise recall should be considered alongside overall accuracy. 
Averaged across models, accuracy increases from 75.4\% at $\tau=0.1$ to 88.4\% at $\tau=0.9$, while a shared threshold of 0.7 achieves 87.4\%, only 2.2 points below the post hoc model-specific optimum. 
However, higher thresholds substantially delay first detection, sometimes by hundreds or more than one thousand tokens. 


\subsection{Dynamic Layer-Selection Threshold}

\begin{table*}[h]
\centering
\small
\setlength{\tabcolsep}{4pt}
\renewcommand{\arraystretch}{1.15}
\caption{Number of dynamically selected layers at different layer-selection thresholds.}
\label{tab:layer_selection_ablation}
\begin{tabular}{ccccccc}
\toprule
\textbf{Selection threshold}  & \textbf{Evaluated $n$} &  \textbf{Mean layers} &  \textbf{Mean / 8} \\
\midrule
\rowcolor[HTML]{F3F4F6} 0.4  & 80  & 1.095  & 13.7\% \\
0.6  & 80 & 1.527 & 19.1\% \\
\rowcolor[HTML]{F3F4F6} 0.7  & 80  & 1.838 & 23.0\% \\
\bottomrule
\end{tabular}
\end{table*}

\begin{table*}[ht]
\centering
\small
\setlength{\tabcolsep}{8pt}
\renewcommand{\arraystretch}{1.15}
\caption{Mean per-layer selection frequency at different layer-selection thresholds.}
\label{tab:layer_selection_frequency}
\begin{tabular}{cccc}
\toprule
\textbf{Layer index} & \textbf{$\tau_{\mathrm{layer}}=0.4$} & \textbf{$\tau_{\mathrm{layer}}=0.6$} & \textbf{$\tau_{\mathrm{layer}}=0.7$} \\
\midrule
\rowcolor[HTML]{F3F4F6} 11 & 15.6\% & 20.3\% & 24.4\% \\
12 & 10.6\% & 15.9\% & 18.3\% \\
\rowcolor[HTML]{F3F4F6} 13 & 15.3\% & 19.4\% & 24.7\% \\
14 & 15.0\% & 20.1\% & 24.4\% \\
\rowcolor[HTML]{F3F4F6} 15 & 14.5\% & 20.9\% & 26.1\% \\
16 & 13.4\% & 19.9\% & 23.4\% \\
\rowcolor[HTML]{F3F4F6} 17 & 13.8\% & 19.6\% & 23.1\% \\
18 & 11.3\% & 16.4\% & 19.4\% \\
\bottomrule
\end{tabular}
\end{table*}

Table~\ref{tab:layer_selection_ablation} reports the overall sensitivity of dynamic layer selection while holding the classifier threshold fixed at 0.5.
Because Equation~\ref{eq:radar_d_dynamic_layers} selects layers satisfying $M_{\mathcal C}^{(l)}(s)<\kappa_{\mathcal C}$, increasing $\kappa_{\mathcal C}$ relaxes the eligibility criterion and therefore expands the intervention set.
Accordingly, raising the selection threshold from 0.4 to 0.7 increases the mean number of selected layers per evaluated step from 1.095 to 1.838, or from 13.7\% to 23.0\% of the eight candidate layers.
The median also rises from 0.021 to 0.514 layers, indicating that selection remains sparse for many trajectories even as the average intervention coverage grows.
Table~\ref{tab:layer_selection_frequency} reports the corresponding per-layer frequencies and shows the same monotonic increase for every candidate layer.
At $\kappa_{\mathcal C}=0.7$, selection frequencies range from 18.3\% for layer 12 to 26.1\% for layer 15, with no single layer dominating the dynamic set.

\subsection{Retained-Layer Fraction}

\begin{table*}[t]
\centering
\scriptsize
\setlength{\tabcolsep}{1.5pt}
\renewcommand{\arraystretch}{1.15}
\caption{Sensitivity of layer localization to the retained-layer fraction $\xi$. Each sweep cell reports the selected layer indices after retaining the $\lceil\xi L\rceil$ highest-PAS candidate layers.}
\label{tab:static_layer_fraction_ablation}
\begin{tabular}{lccccccc}
\toprule
\textbf{Model} & \textbf{$\xi=0.15$} & \textbf{$\xi=0.20$} & \textbf{$\xi=0.25$} & \textbf{$\xi=0.30$} & \textbf{$\xi=0.35$} & \textbf{$\xi=0.40$} & \textbf{Shared subset} \\
\midrule
\rowcolor[HTML]{F3F4F6} Llama-8B & $11$--$15$ & $10$--$15$ & $9$--$16$ & $9$--$16$ & $9$--$16$ & $9$--$16$ & $11$--$15$ \\
Qwen-14B & $25$--$31$ & $22$--$31$ & $22$--$32$ & $22$--$32$ & $21$--$32$ & $21$--$32$ & $25$--$31$ \\
\rowcolor[HTML]{F3F4F6} QwQ-32B & $37$--$46$ & $36$--$48$ & $34$--$49$ & $34$--$49$ & $34$--$49$ & $34$--$49$ & $37$--$46$ \\
Qwen3.6-27B & $\{11,15\}$ & $\{11,15,19\}$ & $\{11,15,19,23\}$ & $\{11,15,19,23\}$ & $\{11,15,19,23\}$ & $\{11,15,19,23\}$ & $\{11,15\}$ \\
\rowcolor[HTML]{F3F4F6} GLM-4.7 & $18$--$24$ & $18$--$26$ & $18$--$27$ & $18$--$27$ & $17$--$27$ & $17$--$27$ & $18$--$24$ \\
\bottomrule
\end{tabular}
\end{table*}

We test whether the static localization is sensitive to the size of the candidate pool by varying $\xi$ from 0.15 to 0.40.
As shown in Table~\ref{tab:static_layer_fraction_ablation}, the selected sets are nested for every model: increasing $\xi$ adds layers but never removes a layer selected at a smaller value.
From $\xi=0.25$ onward, the selected set remains unchanged for Llama-8B, QwQ-32B, and Qwen3.6-27B; Qwen-14B and GLM-4.7 each add only one lower-index boundary layer at $\xi=0.35$.
The shared subsets across all six settings therefore identify a stable model-specific core, consistent with the highest-PAS evidence remaining concentrated in the same layer region as the candidate pool expands.
This sweep establishes robustness of layer localization to $\xi$.

\subsection{Hyperparameter Settings}

Unless otherwise stated, the main RADAR experiments use a classifier confidence
threshold of $\tau=0.5$, a retained-layer fraction of $\xi=0.3$, and a dynamic
layer-selection threshold of $\kappa_{\mathcal C}=0.5$. The minimum online
detection checkpoint is a sequence length of 256 tokens, after which state
inference follows the geometrically spaced schedule defined in
Section~\ref{sec:method_radar_d}. PAS and CTS histories are sampled every 32
decoding steps, while dynamic layer eligibility is updated every 8 steps. Once
the harmful-state gate is activated, the intervention remains active until the
reasoning process terminates.

Trajectory collection uses a sampling temperature of 0.5. In the defense
evaluation, benign requests retain temperature 0.5, whereas attack requests use
temperature 0. All runs use a maximum generation budget of 16k tokens. For the compared external defenses, AUSteer uses $k=16$
selected atomic units with steering strength $\alpha=10$, and the n-gram
baseline blocks repeated trigrams.

\end{document}